\pdfoutput=1

\documentclass[11pt]{article}

\usepackage[preprint]{acl}

\usepackage{times}
\usepackage{latexsym}
\usepackage{microtype}
\usepackage{graphicx}
\usepackage{blindtext} 
\usepackage{subcaption}
\usepackage{booktabs} 
\usepackage{amsmath} 
\usepackage{bbm} 
\usepackage{multicol}
\usepackage{multirow}
\usepackage{booktabs}
\usepackage{caption}
\usepackage{multirow,bigdelim}
\usepackage{makecell}
\usepackage{verbatim}
\usepackage{float}

\usepackage[T1]{fontenc}

\usepackage[utf8]{inputenc}

\usepackage{microtype}

\usepackage{inconsolata}

\usepackage{graphicx}
\usepackage{array}
\PassOptionsToPackage{table}{xcolor}
\usepackage{xcolor}

\newcolumntype{M}[1]{>{\centering\arraybackslash}m{#1}}
\definecolor{lightgray}{gray}{0.9}

\usepackage[most]{tcolorbox}
\usepackage{alltt}
\tcbset{
  aibox/.style={
    width=474.18663pt,
    top=10pt,
    colback=white,
    colframe=black,
    colbacktitle=black,
    enhanced,
    center,
    attach boxed title to top left={yshift=-0.1in,xshift=0.15in},
    boxed title style={boxrule=0pt,colframe=white,},
  }
}
\newtcolorbox{AIbox}[2][]{aibox,title=#2,#1}

\title{Does Prompt Formatting Have Any Impact on LLM Performance?}

\author{
  Jia He$^{1}$\thanks{Equal Contribution}, Mukund Rungta$^{1}$\footnotemark[1], David Koleczek$^1$, Arshdeep Sekhon$^1$,\\
  \textbf{Franklin X Wang$^2$, Sadid Hasan$^1$}\\
  $^1$Microsoft,~$^2$MIT \\
  \texttt{\{hejia, rungtamukund, dkoleczek, asekhon sadidhasan\}@microsoft.com} \\
  \texttt{\{fxwang\}@mit.edu} \\
}  

\begin{document}
\maketitle

\begin{abstract}
In the realm of Large Language Models (LLMs), prompt optimization is crucial for model performance. Although previous research has explored aspects like rephrasing prompt contexts, using various prompting techniques (like in-context learning and chain-of-thought), and ordering few-shot examples, our understanding of LLM sensitivity to prompt templates remains limited. Therefore, this paper examines the impact of different prompt templates on LLM performance. We formatted the same contexts into various human-readable templates, including plain text, Markdown, JSON, and YAML, and evaluated their impact across tasks like natural language reasoning, code generation, and translation using OpenAI’s GPT models. Experiments show that GPT-3.5-turbo’s performance varies by up to 40\% in a code translation task depending on the prompt template, while larger models like GPT-4 are more robust to these variations. Our analysis highlights the need to reconsider the use of fixed prompt templates, as different formats can significantly affect model performance.
\end{abstract}

\section{Introduction}
\label{Introduction}

The emergence of LLMs marks a significant advancement in AI, revolutionizing natural language processing, understanding, and generation (\cite{brown2020language, ouyang2022training, chowdhery2022palm, achiam2023gpt}). Prompt engineering has become crucial, focusing on crafting inputs that guide LLMs to produce desired outputs, leveraging a nuanced understanding of how these models interpret and respond to prompts (\cite{sahoo2024systematic}). Effective prompt design generally includes clear instructions, Retrieval-Augmented Generation (RAG) or other prompting approaches for enhancing in-context learning (ICL), and appropriate formatting.

Often overlooked, prompt format can significantly impact model performance, contrary to the assumption that it remains stable across different templates. There exist limited research and anecdotal evidence (\cite{armentweet, sclar2023quantifying, voronov2024mind}), which suggest that prompt format choices may lead to substantial performance variations, raising concerns about current evaluation standards that ignore this factor. For example, one study showed that LLMs are sensitive to minor fine-grained prompt modifications, such as separators or capitalization changes (\cite{sclar2023quantifying}). Also, existing evaluation approaches typically use fixed templates, potentially leading to misleading conclusions (\cite{voronov2024mind}).

\begin{figure}[!t]
\center
  \includegraphics[width=\columnwidth]{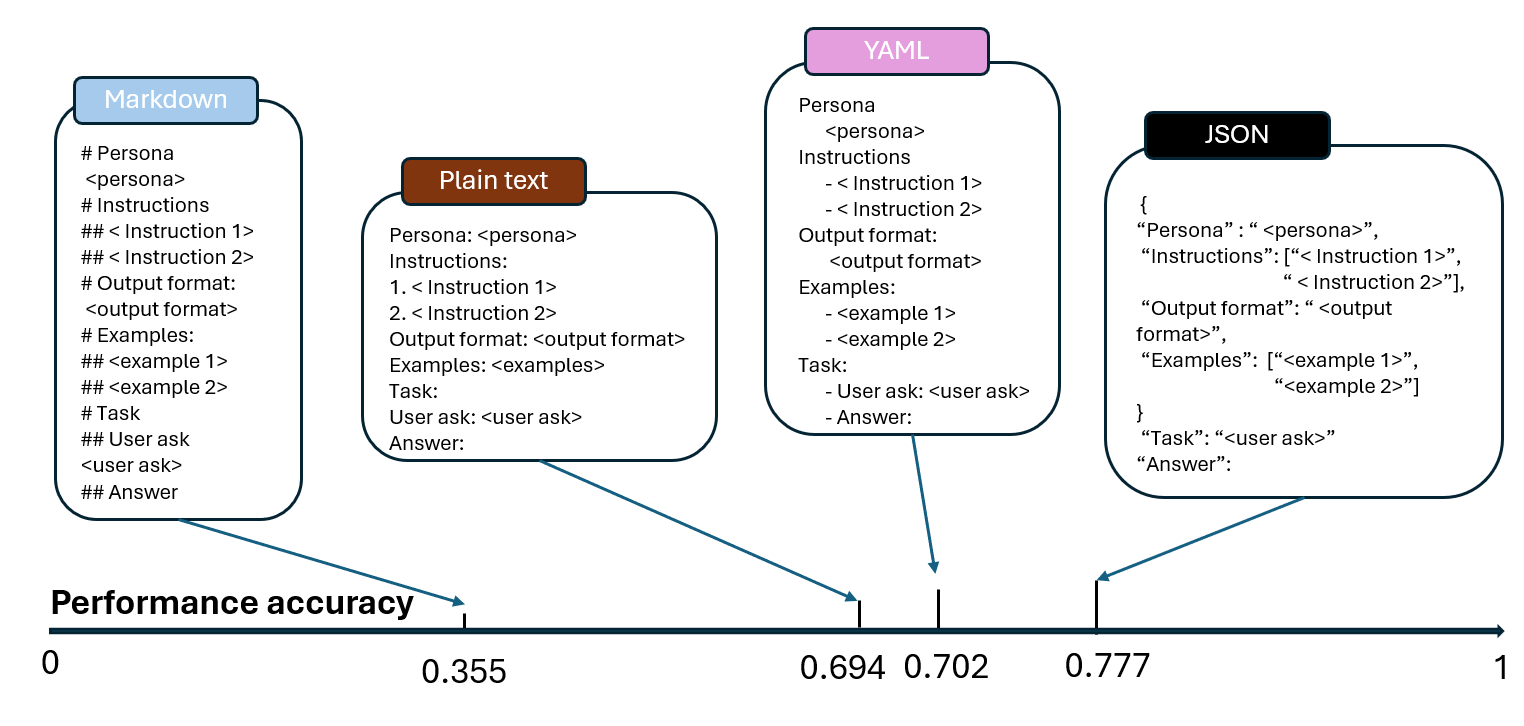}
\caption{An example to demonstrate how prompt formatting impacts GPT-35-turbo-16k-0613 model's performance based on our experiments on  multiple choice questions related to international law  from the MMLU benchmark (\cite{hendrycks2020measuring}). Texts inside "$<>$"  are replaced by actual contexts. Accuracy goes up by 42\% for JSON compared to Markdown.}
\label{fig:pitch}
\end{figure}

Inspired by these findings, our study investigates whether broader changes in prompt format affect model efficacy. We evaluate the impact of prompt templates on OpenAI's four GPT models across six benchmarks, using plain text, Markdown, YAML, and JSON formats, as illustrated in Figure \ref{fig:pitch}. This comprehensive approach contrasts with prior research that primarily examined minor template alterations. Our research focuses on the GPT model series for two main reasons: the lack of comparative analyses of behavioral patterns across different GPT model iterations, especially the latest GPT-4-turbo, and the need to identify effective interaction methods and optimal input formats for these models, which do not disclose their training methodologies or data.



Our study is designed to investigate the following key questions:
\begin{itemize}
\item  \textbf{Sensitivity}: To what extent does the performance of GPT models vary with different prompt formats?
\item  \textbf{Consistency}: Are GPT models capable of producing uniform responses to identical queries when presented with varying prompt structures?
\item  \textbf{Transferability}: Is there an optimal prompt format that is universally effective across diverse GPT models, thereby ensuring peak performance?
\end{itemize}

In addition to our primary questions, we explore the correlation between prompt format efficacy and task-specific competencies, as well as the impact of model size on performance. OpenAI's GPT models including GPT-35-turbo  and GPT-4 \cite{achiam2023gpt} show unpredictable sensitivity to prompt format changes, with significant performance discrepancies across all models and benchmarks. Notably, there is no universally optimal format, even within the same generational lineage. However, GPT-4-turbo demonstrates greater resilience to prompt format changes compared to its predecessors and contemporaries. In summary, our key contributions are as follows:

\begin{itemize}

    \item This study is the first to compare the impact of different prompt formats on GPT models' performance across various tasks, examining plain text, Markdown, YAML, and JSON.
    \item Our research provides an extensive analysis of prompt formatting effects on GPT models across a wide range of tasks, including multiple-choice questions, code generation, and translation.
    \item We present an evaluation of the GPT model iterations via Azure OpenAI, revealing that GPT-4-turbo is less susceptible to prompt structure variations compared to earlier models.
\end{itemize}
\section{Experimental Setup}
\subsection{Datasets}
Our experiments span various tasks and datasets, categorized into three main groups:
\begin{itemize}
    \item \textbf{Natural Language to Natural Language (NL2NL)}: Includes \textbf{Massive Multitask Language Understanding (MMLU)} \cite{hendrycks2020measuring} and \textbf{NER Finance} from OpenAI Evals \cite{openai_evals}.
    \item \textbf{Natural Language to Code (NL2Code)}: Includes \textbf{HumanEval} \cite{chen2021codex} and \textbf{FIND} \cite{schwettmann2023find}.
    \item \textbf{Code to Code (Code2Code)}: Includes \textbf{CODEXGLUE} \cite{lu2021codexglue} and \textbf{HumanEval-X} \cite{zheng2023codegeex}.
\end{itemize}

We initially assess model performance using task-specific scalar scoring functions, followed by metrics from Sections \ref{sec:sensitivity} to \ref{sec:transferability} to address our research questions. Detailed dataset descriptions and metrics are in Appendix \ref{appendix:datadescription}.

\subsection{Prompt Design}
We use various input formats: plain text, markdown, YAML, and JSON. Prompts include five components: \textit{persona}, \textit{task instructions}, \textit{examples}, \textit{output format instructions}, and \textit{user ask}. We ensure the content of each placeholder stays the same across different prompt formats. The only differences are in structure and syntax. To avoid confounding variables, we design the prompts so that the context and meaning remain consistent, regardless of the format. Examples are in Appendix \ref{appendix:prompt_template}.

\subsection{Models}
Experiments were conducted on OpenAI's GPT-3.5 and GPT-4 models via Azure \cite{azure_openai_models}. For GPT-3.5, we used ``gpt-35-turbo-0613'' and ``gpt-35-turbo-16k-0613'' to compare context window sizes (4k vs. 16k). For GPT-4, we used ``gpt-4-32k-0613'' and ``gpt-4-1106-preview'' to test the newer, faster variant with a 128k context window.
\section{Sensitivity}
\label{sec:sensitivity}

\begin{table*}[t]
\centering
\resizebox{\textwidth}{!}{
\small
\begin{tabular}{lcccc c cccc c cccc c cccc}
\hline \addlinespace
 & \multicolumn{3}{c}{GPT-35-turbo-0613} &  & \multicolumn{3}{c}{GPT-35-turbo-16k-0613} &  & \multicolumn{3}{c}{GPT-4-1106-preview} &  & \multicolumn{3}{c}{GPT-4-32k-0613} \\
 \cmidrule(lr){2-4} \cmidrule(lr){6-8} \cmidrule(lr){10-12} \cmidrule(lr){14-16}
& Max & Min & p-value && Max & Min & p-value && Max & Min & p-value && Max & Min & p-value \\
\hline \addlinespace
MMLU &  \makecell[l]{ 59.7 \\ (JSON) } & \makecell[l]{ 50.0\\ (Markdown)} & $< 0.001$  && \makecell[l]{ 59.4 \\ (JSON) } & \makecell[l]{ 50.7\\ (Markdown)} & $< 0.001$ && \makecell[l]{ 81.2 \\ (Markdown) } & \makecell[l]{ 73.9\\ (JSON)} & $< 0.001$ && \makecell[l]{ 81.3 \\ (Markdown) } & \makecell[l]{ 77.8\\ (JSON)} & $< 0.001$\\

\addlinespace \hline \addlinespace

HumanEval & \makecell[l]{ 59.8 \\ (JSON) } & \makecell[l]{ 40.2 \\ (Plain text) } & $< 0.001$ && \makecell[l]{ 57.9 \\ (JSON) } & \makecell[l]{ 37.20 \\ (Plain text) } & $< 0.001$ && \makecell[l]{ 86.6 \\ (Markdown) } & \makecell[l]{ 82.9 \\ (Plain text) } & $0.055$ && \makecell[l]{ 76.2 \\ (Plain text) } & \makecell[l]{ 21.95 \\ (JSON) } & $< 0.001$  \\

\addlinespace \hline \addlinespace

NER Finance & \makecell[l]{ 37.2 \\ (Plain text) } & \makecell[l]{ 24.6 \\ (YAML) } & $< 0.001$ && \makecell[l]{ 36.80 \\ (Plain text) } & \makecell[l]{ 21.8 \\ (YAML) } & $< 0.001$ && \makecell[l]{ 53.8 \\ (YAML) } & \makecell[l]{ 49.3 \\ (Plain text) } & $0.001$ && \makecell[l]{ 53.2 \\ (YAML) } & \makecell[l]{ 47.2 \\ (Plain text) } & $< 0.001$ \\ 
\addlinespace \hline \addlinespace

\makecell[l]{ CODEXGLUE \\(Java2CS)  } & \makecell[l]{ 78.4 \\ (JSON) } & \makecell[l]{ 66.5 \\ (Plain text) } & $< 0.001$ && \makecell[l]{ 78.4 \\ (JSON) } & \makecell[l]{ 66.5 \\ (Plain text) } & $< 0.001$ && \makecell[l]{ 77.0 \\ (Markdown) } & \makecell[l]{ 68.2 \\ (Plain text) } & $< 0.001$ && \makecell[l]{ 74.2 \\ (Markdown) } & \makecell[l]{ 67.2 \\ (Plain text) } & $< 0.001$ \\

\addlinespace \hline \addlinespace

\makecell[l]{ CODEXGLUE \\(Cs2Java)  } & \makecell[l]{ 77.5 \\ (JSON) } & \makecell[l]{ 68.8 \\ (Plain text) } & $< 0.001$ && \makecell[l]{ 77.5 \\ (JSON) } & \makecell[l]{ 68.9 \\ (Plain text) } & $< 0.001$ && \makecell[l]{ 83.1 \\ (JSON) } & \makecell[l]{ 68.1 \\ (Plain text) } & $< 0.001$ && \makecell[l]{ 75.0 \\ (JSON) } & \makecell[l]{ 67.9 \\ (Plain text) } & $< 0.001$ \\

\addlinespace \hline \addlinespace

HumanEval-X & \makecell[l]{ 69.8 \\ (YAML) } & \makecell[l]{ 63.0 \\ (Plain text) } & $< 0.001$ && \makecell[l]{ 69.8 \\ (YAML) } & \makecell[l]{ 62.9 \\ (Plain text) } & $< 0.001$ && \makecell[l]{ 72.4 \\ (JSON) } & \makecell[l]{ 63.9 \\ (Plain text) } & $< 0.001$ && \makecell[l]{ 72.3 \\ (JSON) } & \makecell[l]{ 65.0 \\ (Plain text) } & $< 0.001$ \\

\addlinespace \hline \addlinespace
FIND & \makecell[l]{ 15.9 \\ (Plain text) } & \makecell[l]{ 5.2 \\ (Markdown) } & $< 0.001$ && \makecell[l]{ 15.8 \\ (Plain text) } & \makecell[l]{  5.0 \\ (Markdown) } & $< 0.001$ && \makecell[l]{  20.7 \\ (Markdown) } & \makecell[l]{ 20.08 \\ (Plain text) } & $< 0.0269$ && \makecell[l]{  21.9 \\ (plaintext) } & \makecell[l]{  17.4 \\ (markdown) } & $< 0.001$ \\

\addlinespace \hline \addlinespace
\end{tabular}
}
\caption{Sensitivity of model performance to prompt format assessed using one-sided matched pair t-tests. Table displays metrics for top and bottom formats (Max/Min) and p-values for each dataset/model. All p-values are below 0.05, except for GPT-4-1104-preview on HumanEval, confirming widespread prompt format sensitivity.}
\label{table:sensitivity}
\end{table*}

\subsection{Metrics Definition}

\paragraph{Sensitivity.}To evaluate how much the choice of prompt template impacts a model's performance on a task \( \mathbf{T} \), we look at a variety of templates \( \{p_1, p_2, \ldots, p_n\} \) and measure their performance using a scoring function \( s \). We identify the highest score \( Max(s(p_i, \mathbf{T})) \) and the lowest score \( Min(s(p_i, \mathbf{T})) \) achieved by the templates for task \( \mathbf{T} \). Then, we use a matched pairs t-test to determine if the performance difference is statistically significant. If the test shows significance, the prompt format matters; if not, the model's performance is relatively insensitive to the prompt used.

\subsection{Does prompt format impact the performance of language models and how significant is the performance variation when switching prompt formats?}

We begin by analyzing if model performance is sensitive to any changes in the prompt format at all. To assess this, we conducted a one-sided matched pair t-test, comparing the best and worst performing formats for each model across various benchmarks. The resulting p-values, which are shown in Table \ref{table:sensitivity}, are mostly below 0.01. This suggests that the differences in model performance due to format changes are statistically significant.

Figure \ref{fig:dotplot} visualizes how the models fare across all benchmarks, highlighting a considerable range in performance. For instance, in the FIND dataset, both GPT-35-turbo-0613 and GPT-35-turbo-16k-0613 show a dramatic $200\%$ improvement when prompts are switched from Markdown to plain text. Similarly, for the HumanEval benchmark, the GPT-4 model with a 32k-0613 configuration exhibits an impressive performance boost of over $300\%$ when the prompt format is changed from JSON to plain text. \textbf{This suggests, LLM performance may not be robust to the choice of prompt format.}


\section{Consistency}

\subsection{Metrics Definition}
Following the sensitivity measurement, we quantify the extent of answer variation due to prompt changes using the \textit{consistency} metric from \cite{shu2023you}. This metric calculates the proportion of test samples that yield identical responses for two prompt templates. The consistency \( C(P_a, P_b) \) for templates \( P_a \) and \( P_b \) is defined as:
$$ C(P_a, P_b) = \frac{1}{N} \sum_{i=1}^{N} \mathbbm{1}\left(A_{P_a}(x_i) = A_{P_b}(x_i)\right) $$
where \( N \) is the test set size and \( A \) represents the model's answer. A higher score indicates greater answer consistency between prompts.

\begin{figure}[hbt!]
  \centering
  \begin{subfigure}{.23\textwidth}
    \centering
    \includegraphics[width=\linewidth]{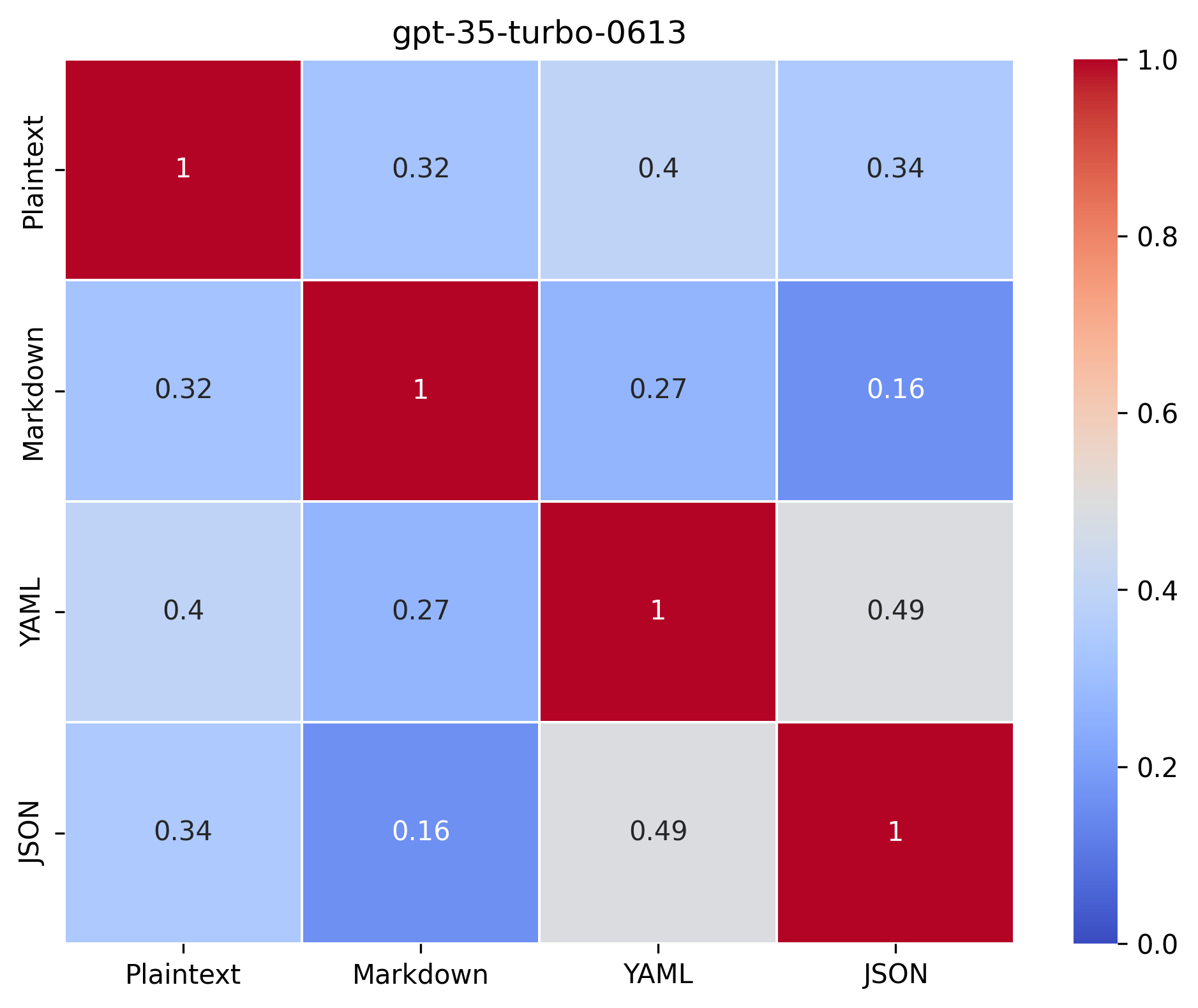}
    \caption{\fontsize{6}{8}\selectfont GPT-35-Turbo-0613}
    \label{fig:consistency-mmlu-gpt35}
  \end{subfigure}%
  \hfill
  \begin{subfigure}{.23\textwidth}
    \centering
    \includegraphics[width=\linewidth]{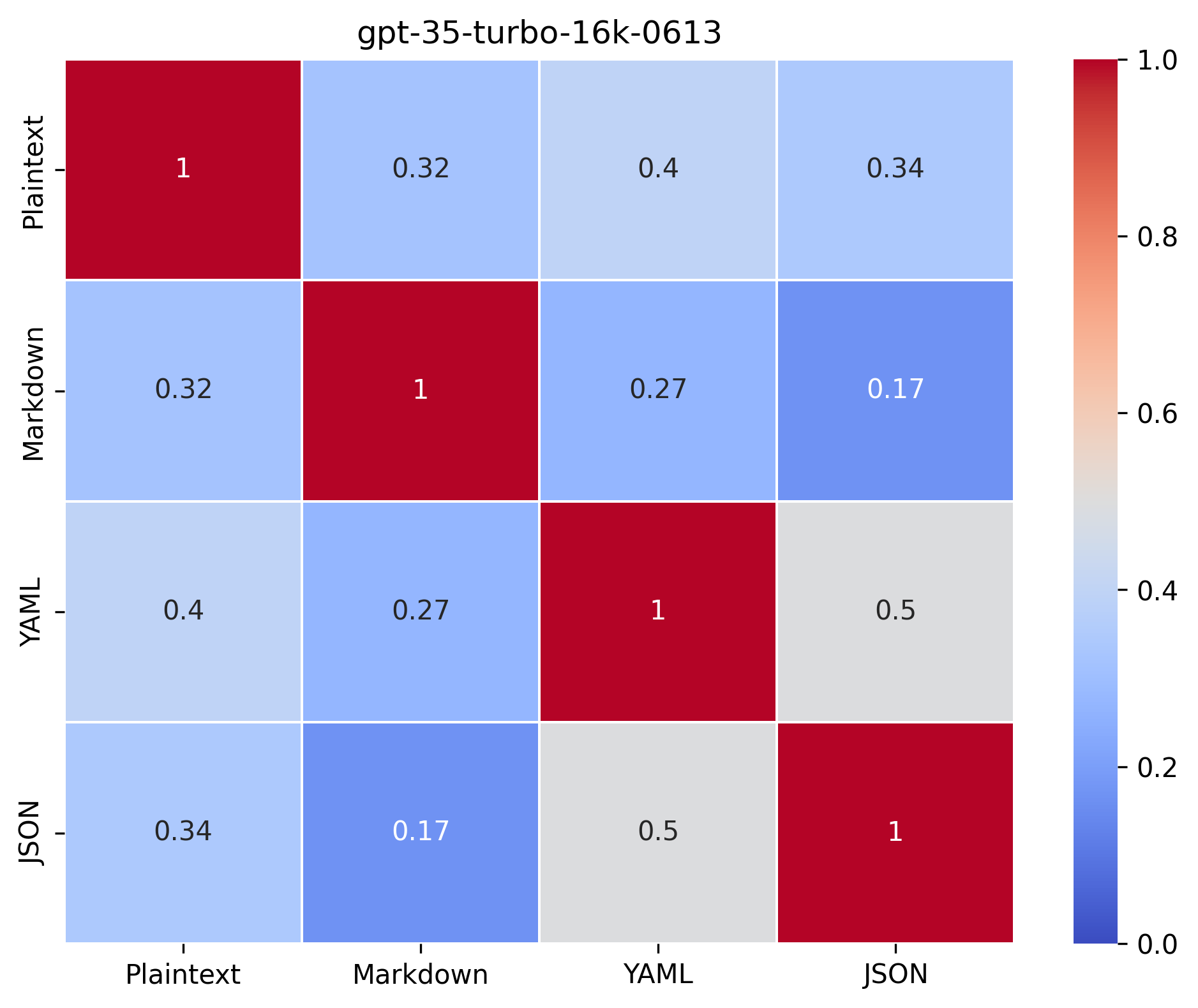}
    \caption{\fontsize{6}{8}\selectfont GPT-35-turbo-16k-0613}
    \label{fig:consistency-mmlu-gpt35-16k}
  \end{subfigure}%
  \hfill
  \begin{subfigure}{.23\textwidth}
    \centering
    \includegraphics[width=\linewidth]{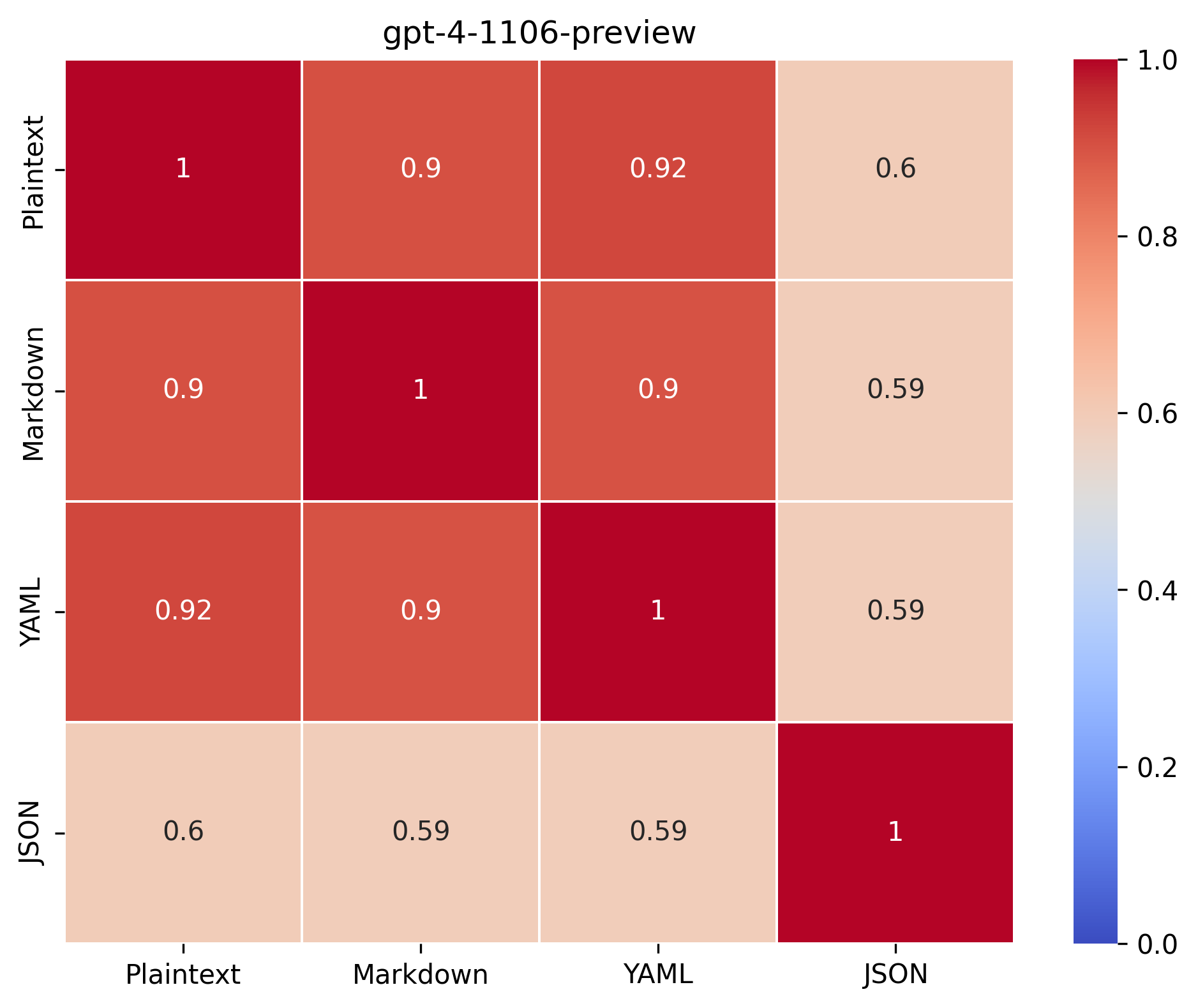}
    \caption{\fontsize{6}{8}\selectfont GPT-4-1106-preview}
    \label{fig:consistency-mmlu-gpt4}
  \end{subfigure}%
  \hfill
  \begin{subfigure}{.23\textwidth}
    \centering
    \includegraphics[width=\linewidth]{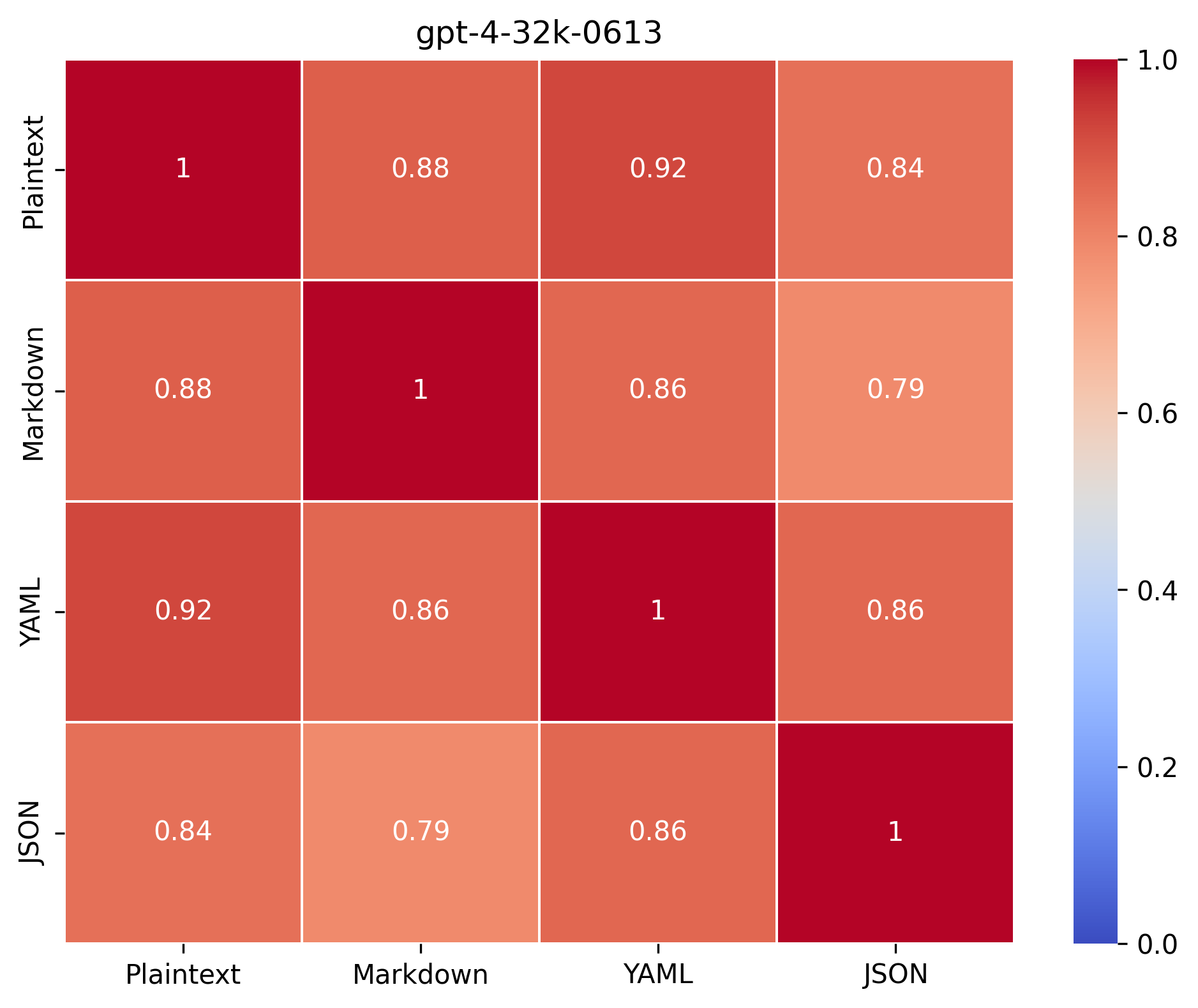}
    \caption{\fontsize{6}{8}\selectfont GPT-4-32k-0613}
    \label{fig:consistency-mmlu-gpt4-32k}
  \end{subfigure}
  \caption{Consistency comparison for MMLU dataset: GPT-3.5 models show consistency scores below 0.5 across format pairs, whereas GPT-4 consistently exceeds 0.5, indicating greater reliability. }
  \label{fig:consistency-mmlu}
\end{figure}

\subsection{Are larger models more consistent in generated outputs between templates?}

Our study assessed the consistency of model outputs using the MMLU and FIND datasets, as shown in Figures \ref{fig:consistency-mmlu} and \ref{fig:consistency-find}. For MMLU, we set the temperature to zero to eliminate response variability. The GPT-3.5-turbo series displayed low consistency, with scores below 0.5, and only 16\% identical responses between Markdown and JSON formats. In contrast, GPT-4's consistency scores surpassed 0.5, indicating better reliability across different prompts. For the FIND dataset, following the settings from \cite{schwettmann2023find}, GPT-4 again outperformed the GPT-3.5-turbo series in consistency. These findings suggest that larger models like GPT-4 are more consistent, but there is still a need for model improvements to achieve reliable performance across various formats. \textbf{In summary, the consistency of model responses varies with size, with larger models like GPT-4 providing more uniform outputs across different prompts.}

\section{Transferability}
\label{sec:transferability}
\subsection{Metrics Definition}

\paragraph{Intersection-over-Union.} To assess the transferability of prompt templates between models, we calculate the Intersection-over-Union (IoU) for the sets of top-performing templates between model pairs. Top-performing templates are those with statistically indistinguishable performance, determined by a matched pairs t-test. The IoU is defined as:
$$\text{IoU} = \frac{|P_{m1} \cap P_{m2}|}{|P_{m1} \cup P_{m2}|}$$
where $P_{m1}$ and $P_{m2}$ represent the sets of top templates for models $m1$ and $m2$, respectively. An IoU threshold of 0.5 is common, but a higher threshold like 0.7 indicates greater overlap.

\subsection{Do models from same family exhibit similar trend across prompt formats?}

\begin{figure}[!t]
\center
  \includegraphics[width=0.4\textwidth]{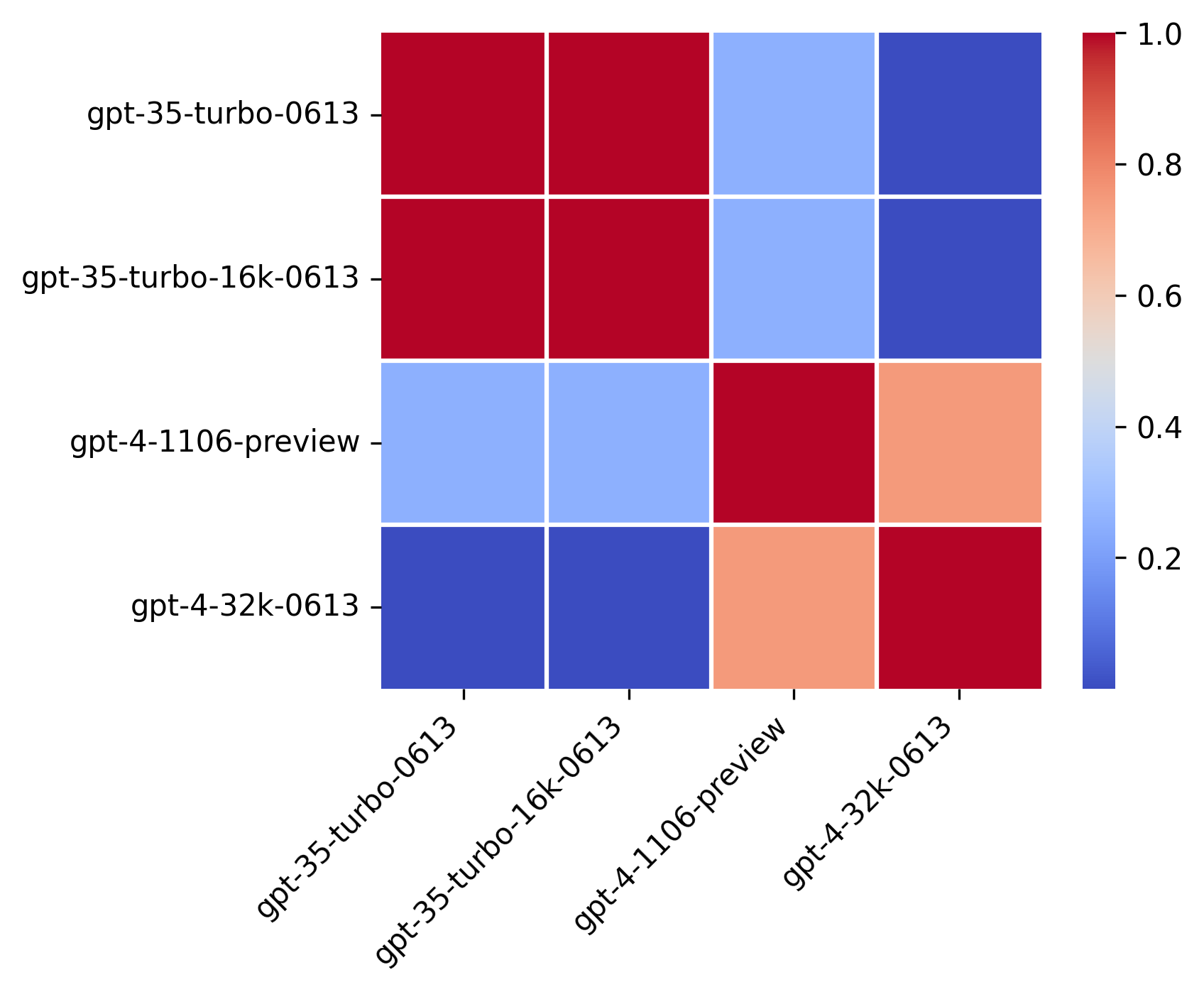}
\caption{Intersection over Union (IoU) scores for top templates on the NER Finance benchmark across models. Higher IoU is observed within same-version model pairs, whereas cross-version pairs exhibit lower IoU.}
\label{fig:mmluiou}
\end{figure}

Our research into Large Language Models (LLMs), GPT-based models in particular, reveals that prompt formatting preferences vary by model. As demonstrated in Figure \ref{fig:mmlubreakdowndotplot}, GPT-3.5-turbo prefers JSON, whereas GPT-4 favors Markdown. When examining prompt transferability using Intersection-over-Union (IoU) metrics (Figure \ref{fig:mmluiou} and Appendix B), we found low compatibility between different model series, with IoU often below 0.2. However, models from the same sub-series, like GPT-35-turbo-16k-0613 and GPT-35-turbo-0613, show high IoU over 0.7.

These insights highlight that even with common architectures and training goals, GPT-models react differently to identical prompts. Optimal performance requires model-specific prompt engineering, as no single format works universally across various GPT models, even within the same family. \textbf{This underscores the necessity for tailored prompt engineering due to the non-transferability of prompt formats across different GPT models.}

\section{Conclusion}
Our study reveals that the way prompts are formatted significantly impacts GPT-based models' performance, with no single format excelling universally. This finding questions current evaluation methods that often ignore prompt structure, potentially misjudging a model's true abilities. We advocate for diverse prompt formats in future LLM testing to accurately gauge and enhance their performance.

Regarding explainability, we observe that model size affects model's responses to prompt variations. For instance, GPT-4's performance is less influenced by prompt changes compared to GPT-3.5, suggesting that larger models may process prompts more consistently. This discovery prompts further research into LLM interpretability, aiming to refine AI adaptability and human-AI interaction.
\section{Limitations}
\label{Limitations}
This study was focused on GPT-based models, however, we plan to examine the impact of prompt formats on other models, such as LLaMA \cite{touvron2023llama}, Gemini \cite{team2023gemini}, PaLM \cite{chowdhery2022palm}, or  smaller models like Phi \cite{li2023textbooks} in the future. This would provide a more holistic understanding of the influence that prompt formatting exerts across different LLM families.

Moreover, there is an opportunity to enhance the breadth of template exploration in subsequent studies. Our research did not include formats like HTML or XML, which are prevalent in the training datasets of many models. Incorporating these formats could yield a more exhaustive examination of prompt format effects.

Lastly, our experimental design maintained all other prompt design elements constant, isolating prompt format as the sole variable. It would be intriguing for future work to investigate how the sensitivity of models to prompt format might shift when other prompt engineering techniques are modified. This includes varying the number of few-shot examples provided or refining the precision of prompt instructions. Such research could offer valuable insights into the interplay between prompt structure and model responsiveness, potentially informing more effective prompt engineering practices.

\bibliography{mybib}

\begin{thebibliography}{34}
\providecommand{\natexlab}[1]{#1}

\bibitem[{Achiam et~al.(2023)Achiam, Adler, Agarwal, Ahmad, Akkaya, Aleman, Almeida, Altenschmidt, Altman, Anadkat et~al.}]{achiam2023gpt}
Josh Achiam, Steven Adler, Sandhini Agarwal, Lama Ahmad, Ilge Akkaya, Florencia~Leoni Aleman, Diogo Almeida, Janko Altenschmidt, Sam Altman, Shyamal Anadkat, et~al. 2023.
\newblock Gpt-4 technical report.
\newblock \emph{arXiv preprint arXiv:2303.08774}.

\bibitem[{Aghajanyan(June 2023)}]{armentweet}
Armen Aghajanyan. June 2023.
\newblock \href {https://twitter.com/ArmenAgha/status/1669084129261162497.} {Tweet: Susan and i found mmlu performance jump 6-10 points in the 40s by formatting multiple choice as (a) not a in mmlu (for internal model). all evaluation of llm’s are broken. evaluating a task requires marginalizing across all prompts that describe the task, not point estimate of one.}

\bibitem[{Brown et~al.(2020)Brown, Mann, Ryder, Subbiah, Kaplan, Dhariwal, Neelakantan, Shyam, Sastry, Askell, Agarwal, Herbert-Voss, Krueger, Henighan, Child, Ramesh, Ziegler, Wu, Winter, Hesse, Chen, Sigler, Litwin, Gray, Chess, Clark, Berner, McCandlish, Radford, Sutskever, and Amodei}]{brown2020language}
Tom~B. Brown, Benjamin Mann, Nick Ryder, Melanie Subbiah, Jared Kaplan, Prafulla Dhariwal, Arvind Neelakantan, Pranav Shyam, Girish Sastry, Amanda Askell, Sandhini Agarwal, Ariel Herbert-Voss, Gretchen Krueger, Tom Henighan, Rewon Child, Aditya Ramesh, Daniel~M. Ziegler, Jeffrey Wu, Clemens Winter, Christopher Hesse, Mark Chen, Eric Sigler, Mateusz Litwin, Scott Gray, Benjamin Chess, Jack Clark, Christopher Berner, Sam McCandlish, Alec Radford, Ilya Sutskever, and Dario Amodei. 2020.
\newblock \href {https://arxiv.org/abs/2005.14165} {Language models are few-shot learners}.
\newblock \emph{Preprint}, arXiv:2005.14165.

\bibitem[{Bubeck et~al.(2023)Bubeck, Chandrasekaran, Eldan, Gehrke, Horvitz, Kamar, Lee, Lee, Li, Lundberg et~al.}]{bubeck2023sparks}
S{\'e}bastien Bubeck, Varun Chandrasekaran, Ronen Eldan, Johannes Gehrke, Eric Horvitz, Ece Kamar, Peter Lee, Yin~Tat Lee, Yuanzhi Li, Scott Lundberg, et~al. 2023.
\newblock Sparks of artificial general intelligence: Early experiments with gpt-4. arxiv.
\newblock \emph{arXiv preprint arXiv:2303.12712}.

\bibitem[{Carlini et~al.(2024)Carlini, Paleka, Dvijotham, Steinke, Hayase, Cooper, Lee, Jagielski, Nasr, Conmy et~al.}]{carlini2024stealing}
Nicholas Carlini, Daniel Paleka, Krishnamurthy~Dj Dvijotham, Thomas Steinke, Jonathan Hayase, A~Feder Cooper, Katherine Lee, Matthew Jagielski, Milad Nasr, Arthur Conmy, et~al. 2024.
\newblock Stealing part of a production language model.
\newblock \emph{arXiv preprint arXiv:2403.06634}.

\bibitem[{Chen et~al.(2021)Chen, Tworek, Jun, Yuan, de~Oliveira~Pinto, Kaplan, Edwards, Burda, Joseph, Brockman, Ray, Puri, Krueger, Petrov, Khlaaf, Sastry, Mishkin, Chan, Gray, Ryder, Pavlov, Power, Kaiser, Bavarian, Winter, Tillet, Such, Cummings, Plappert, Chantzis, Barnes, Herbert-Voss, Guss, Nichol, Paino, Tezak, Tang, Babuschkin, Balaji, Jain, Saunders, Hesse, Carr, Leike, Achiam, Misra, Morikawa, Radford, Knight, Brundage, Murati, Mayer, Welinder, McGrew, Amodei, McCandlish, Sutskever, and Zaremba}]{chen2021codex}
Mark Chen, Jerry Tworek, Heewoo Jun, Qiming Yuan, Henrique~Ponde de~Oliveira~Pinto, Jared Kaplan, Harri Edwards, Yuri Burda, Nicholas Joseph, Greg Brockman, Alex Ray, Raul Puri, Gretchen Krueger, Michael Petrov, Heidy Khlaaf, Girish Sastry, Pamela Mishkin, Brooke Chan, Scott Gray, Nick Ryder, Mikhail Pavlov, Alethea Power, Lukasz Kaiser, Mohammad Bavarian, Clemens Winter, Philippe Tillet, Felipe~Petroski Such, Dave Cummings, Matthias Plappert, Fotios Chantzis, Elizabeth Barnes, Ariel Herbert-Voss, William~Hebgen Guss, Alex Nichol, Alex Paino, Nikolas Tezak, Jie Tang, Igor Babuschkin, Suchir Balaji, Shantanu Jain, William Saunders, Christopher Hesse, Andrew~N. Carr, Jan Leike, Josh Achiam, Vedant Misra, Evan Morikawa, Alec Radford, Matthew Knight, Miles Brundage, Mira Murati, Katie Mayer, Peter Welinder, Bob McGrew, Dario Amodei, Sam McCandlish, Ilya Sutskever, and Wojciech Zaremba. 2021.
\newblock \href {https://arxiv.org/abs/2107.03374} {Evaluating large language models trained on code}.

\bibitem[{Chowdhery et~al.(2022)Chowdhery, Narang, Devlin, Bosma, Mishra, Roberts, Barham, Chung, Sutton, Gehrmann, Schuh, Shi, Tsvyashchenko, Maynez, Rao, Barnes, Tay, Shazeer, Prabhakaran, Reif, Du, Hutchinson, Pope, Bradbury, Austin, Isard, Gur-Ari, Yin, Duke, Levskaya, Ghemawat, Dev, Michalewski, Garcia, Misra, Robinson, Fedus, Zhou, Ippolito, Luan, Lim, Zoph, Spiridonov, Sepassi, Dohan, Agrawal, Omernick, Dai, Pillai, Pellat, Lewkowycz, Moreira, Child, Polozov, Lee, Zhou, Wang, Saeta, Diaz, Firat, Catasta, Wei, Meier-Hellstern, Eck, Dean, Petrov, and Fiedel}]{chowdhery2022palm}
Aakanksha Chowdhery, Sharan Narang, Jacob Devlin, Maarten Bosma, Gaurav Mishra, Adam Roberts, Paul Barham, Hyung~Won Chung, Charles Sutton, Sebastian Gehrmann, Parker Schuh, Kensen Shi, Sasha Tsvyashchenko, Joshua Maynez, Abhishek Rao, Parker Barnes, Yi~Tay, Noam Shazeer, Vinodkumar Prabhakaran, Emily Reif, Nan Du, Ben Hutchinson, Reiner Pope, James Bradbury, Jacob Austin, Michael Isard, Guy Gur-Ari, Pengcheng Yin, Toju Duke, Anselm Levskaya, Sanjay Ghemawat, Sunipa Dev, Henryk Michalewski, Xavier Garcia, Vedant Misra, Kevin Robinson, Liam Fedus, Denny Zhou, Daphne Ippolito, David Luan, Hyeontaek Lim, Barret Zoph, Alexander Spiridonov, Ryan Sepassi, David Dohan, Shivani Agrawal, Mark Omernick, Andrew~M. Dai, Thanumalayan~Sankaranarayana Pillai, Marie Pellat, Aitor Lewkowycz, Erica Moreira, Rewon Child, Oleksandr Polozov, Katherine Lee, Zongwei Zhou, Xuezhi Wang, Brennan Saeta, Mark Diaz, Orhan Firat, Michele Catasta, Jason Wei, Kathy Meier-Hellstern, Douglas Eck, Jeff Dean, Slav Petrov, and Noah Fiedel. 2022.
\newblock \href {https://arxiv.org/abs/2204.02311} {Palm: Scaling language modeling with pathways}.
\newblock \emph{Preprint}, arXiv:2204.02311.

\bibitem[{Hendrycks et~al.(2020)Hendrycks, Burns, Basart, Zou, Mazeika, Song, and Steinhardt}]{hendrycks2020measuring}
Dan Hendrycks, Collin Burns, Steven Basart, Andy Zou, Mantas Mazeika, Dawn Song, and Jacob Steinhardt. 2020.
\newblock Measuring massive multitask language understanding.
\newblock \emph{arXiv preprint arXiv:2009.03300}.

\bibitem[{Lewis et~al.(2021)Lewis, Perez, Piktus, Petroni, Karpukhin, Goyal, Küttler, Lewis, tau Yih, Rocktäschel, Riedel, and Kiela}]{lewis2021retrievalaugmented}
Patrick Lewis, Ethan Perez, Aleksandra Piktus, Fabio Petroni, Vladimir Karpukhin, Naman Goyal, Heinrich Küttler, Mike Lewis, Wen tau Yih, Tim Rocktäschel, Sebastian Riedel, and Douwe Kiela. 2021.
\newblock \href {https://arxiv.org/abs/2005.11401} {Retrieval-augmented generation for knowledge-intensive nlp tasks}.
\newblock \emph{Preprint}, arXiv:2005.11401.

\bibitem[{Li et~al.(2023)Li, Bubeck, Eldan, Del~Giorno, Gunasekar, and Lee}]{li2023textbooks}
Yuanzhi Li, S{\'e}bastien Bubeck, Ronen Eldan, Allie Del~Giorno, Suriya Gunasekar, and Yin~Tat Lee. 2023.
\newblock Textbooks are all you need ii: phi-1.5 technical report.
\newblock \emph{arXiv preprint arXiv:2309.05463}.

\bibitem[{Liu et~al.(2023)Liu, Lin, Hewitt, Paranjape, Bevilacqua, Petroni, and Liang}]{liu2023lost}
Nelson~F. Liu, Kevin Lin, John Hewitt, Ashwin Paranjape, Michele Bevilacqua, Fabio Petroni, and Percy Liang. 2023.
\newblock \href {https://arxiv.org/abs/2307.03172} {Lost in the middle: How language models use long contexts}.
\newblock \emph{Preprint}, arXiv:2307.03172.

\bibitem[{Lu et~al.(2021)Lu, Guo, Ren, Huang, Svyatkovskiy, Blanco, Clement, Drain, Jiang, Tang et~al.}]{lu2021codexglue}
Shuai Lu, Daya Guo, Shuo Ren, Junjie Huang, Alexey Svyatkovskiy, Ambrosio Blanco, Colin Clement, Dawn Drain, Daxin Jiang, Duyu Tang, et~al. 2021.
\newblock Codexglue: A machine learning benchmark dataset for code understanding and generation.
\newblock \emph{arXiv preprint arXiv:2102.04664}.

\bibitem[{Lu et~al.(2022)Lu, Bartolo, Moore, Riedel, and Stenetorp}]{lu2022fantastically}
Yao Lu, Max Bartolo, Alastair Moore, Sebastian Riedel, and Pontus Stenetorp. 2022.
\newblock \href {https://arxiv.org/abs/2104.08786} {Fantastically ordered prompts and where to find them: Overcoming few-shot prompt order sensitivity}.
\newblock \emph{Preprint}, arXiv:2104.08786.

\bibitem[{Microsoft()}]{guidance}
Microsoft.
\newblock guidance.
\newblock https://github.com/guidance-ai.

\bibitem[{Microsoft(2024)}]{azure_openai_models}
Microsoft. 2024.
\newblock Azure openai service models.
\newblock \url{https://learn.microsoft.com/en-us/azure/ai-services/openai/concepts/models#gpt-4-and-gpt-4-turbo-preview}.
\newblock Accessed: 2024-03-26.

\bibitem[{OpenAI(2023)}]{openai_evals}
OpenAI. 2023.
\newblock Evals.
\newblock \url{https://github.com/openai/evals}.

\bibitem[{OpenAI(2024)}]{OpenAI2024}
OpenAI. 2024.
\newblock \href {https://openai.com/blog/new-embedding-models-and-api-updates} {New embedding models and api updates}.
\newblock Accessed: 2024-03-26.

\bibitem[{OpenAI(November 2023)}]{OpenAI}
OpenAI. November 2023.
\newblock \href {https://openai.com/blog/new-models-and-developer-products-announced-at-devday} {Improved instruction following and json mode}.

\bibitem[{Ouyang et~al.(2022)Ouyang, Wu, Jiang, Almeida, Wainwright, Mishkin, Zhang, Agarwal, Slama, Ray, Schulman, Hilton, Kelton, Miller, Simens, Askell, Welinder, Christiano, Leike, and Lowe}]{ouyang2022training}
Long Ouyang, Jeff Wu, Xu~Jiang, Diogo Almeida, Carroll~L. Wainwright, Pamela Mishkin, Chong Zhang, Sandhini Agarwal, Katarina Slama, Alex Ray, John Schulman, Jacob Hilton, Fraser Kelton, Luke Miller, Maddie Simens, Amanda Askell, Peter Welinder, Paul Christiano, Jan Leike, and Ryan Lowe. 2022.
\newblock \href {https://arxiv.org/abs/2203.02155} {Training language models to follow instructions with human feedback}.
\newblock \emph{Preprint}, arXiv:2203.02155.

\bibitem[{Papineni et~al.(2002)Papineni, Roukos, Ward, and Zhu}]{papineni2002bleu}
Kishore Papineni, Salim Roukos, Todd Ward, and Wei-Jing Zhu. 2002.
\newblock Bleu: a method for automatic evaluation of machine translation.
\newblock In \emph{Proceedings of the 40th annual meeting of the Association for Computational Linguistics}, pages 311--318.

\bibitem[{Sahoo et~al.(2024)Sahoo, Singh, Saha, Jain, Mondal, and Chadha}]{sahoo2024systematic}
Pranab Sahoo, Ayush~Kumar Singh, Sriparna Saha, Vinija Jain, Samrat Mondal, and Aman Chadha. 2024.
\newblock \href {https://arxiv.org/abs/2402.07927} {A systematic survey of prompt engineering in large language models: Techniques and applications}.
\newblock \emph{Preprint}, arXiv:2402.07927.

\bibitem[{Schwettmann et~al.(2023)Schwettmann, Shaham, Materzynska, Chowdhury, Li, Andreas, Bau, and Torralba}]{schwettmann2023find}
Sarah Schwettmann, Tamar~Rott Shaham, Joanna Materzynska, Neil Chowdhury, Shuang Li, Jacob Andreas, David Bau, and Antonio Torralba. 2023.
\newblock \href {https://arxiv.org/abs/2309.03886} {Find: A function description benchmark for evaluating interpretability methods}.
\newblock \emph{Preprint}, arXiv:2309.03886.

\bibitem[{Sclar et~al.(2023)Sclar, Choi, Tsvetkov, and Suhr}]{sclar2023quantifying}
Melanie Sclar, Yejin Choi, Yulia Tsvetkov, and Alane Suhr. 2023.
\newblock Quantifying language models' sensitivity to spurious features in prompt design or: How i learned to start worrying about prompt formatting.
\newblock \emph{arXiv preprint arXiv:2310.11324}.

\bibitem[{Shu et~al.(2023)Shu, Zhang, Choi, Dunagan, Card, and Jurgens}]{shu2023you}
Bangzhao Shu, Lechen Zhang, Minje Choi, Lavinia Dunagan, Dallas Card, and David Jurgens. 2023.
\newblock You don't need a personality test to know these models are unreliable: Assessing the reliability of large language models on psychometric instruments.
\newblock \emph{arXiv preprint arXiv:2311.09718}.

\bibitem[{Sui et~al.(2024)Sui, Zhou, Zhou, Han, and Zhang}]{sui2024table}
Yuan Sui, Mengyu Zhou, Mingjie Zhou, Shi Han, and Dongmei Zhang. 2024.
\newblock Table meets llm: Can large language models understand structured table data? a benchmark and empirical study.
\newblock In \emph{The 17th ACM International Conference on Web Search and Data Mining (WSDM '24)}.

\bibitem[{Team et~al.(2023)Team, Anil, Borgeaud, Wu, Alayrac, Yu, Soricut, Schalkwyk, Dai, Hauth et~al.}]{team2023gemini}
Gemini Team, Rohan Anil, Sebastian Borgeaud, Yonghui Wu, Jean-Baptiste Alayrac, Jiahui Yu, Radu Soricut, Johan Schalkwyk, Andrew~M Dai, Anja Hauth, et~al. 2023.
\newblock Gemini: a family of highly capable multimodal models.
\newblock \emph{arXiv preprint arXiv:2312.11805}.

\bibitem[{Touvron et~al.(2023)Touvron, Lavril, Izacard, Martinet, Lachaux, Lacroix, Rozi{\`e}re, Goyal, Hambro, Azhar et~al.}]{touvron2023llama}
Hugo Touvron, Thibaut Lavril, Gautier Izacard, Xavier Martinet, Marie-Anne Lachaux, Timoth{\'e}e Lacroix, Baptiste Rozi{\`e}re, Naman Goyal, Eric Hambro, Faisal Azhar, et~al. 2023.
\newblock Llama: Open and efficient foundation language models.
\newblock \emph{arXiv preprint arXiv:2302.13971}.

\bibitem[{Voronov et~al.(2024)Voronov, Wolf, and Ryabinin}]{voronov2024mind}
Anton Voronov, Lena Wolf, and Max Ryabinin. 2024.
\newblock Mind your format: Towards consistent evaluation of in-context learning improvements.
\newblock \emph{arXiv preprint arXiv:2401.06766}.

\bibitem[{Wang et~al.(2023)Wang, Wei, Schuurmans, Le, Chi, Narang, Chowdhery, and Zhou}]{wang2023selfconsistency}
Xuezhi Wang, Jason Wei, Dale Schuurmans, Quoc Le, Ed~Chi, Sharan Narang, Aakanksha Chowdhery, and Denny Zhou. 2023.
\newblock \href {https://arxiv.org/abs/2203.11171} {Self-consistency improves chain of thought reasoning in language models}.
\newblock \emph{Preprint}, arXiv:2203.11171.

\bibitem[{Wei et~al.(2023)Wei, Wang, Schuurmans, Bosma, Ichter, Xia, Chi, Le, and Zhou}]{wei2023chainofthought}
Jason Wei, Xuezhi Wang, Dale Schuurmans, Maarten Bosma, Brian Ichter, Fei Xia, Ed~Chi, Quoc Le, and Denny Zhou. 2023.
\newblock \href {https://arxiv.org/abs/2201.11903} {Chain-of-thought prompting elicits reasoning in large language models}.
\newblock \emph{Preprint}, arXiv:2201.11903.

\bibitem[{Yao et~al.(2023)Yao, Zhao, Yu, Du, Shafran, Narasimhan, and Cao}]{yao2023react}
Shunyu Yao, Jeffrey Zhao, Dian Yu, Nan Du, Izhak Shafran, Karthik Narasimhan, and Yuan Cao. 2023.
\newblock \href {https://arxiv.org/abs/2210.03629} {React: Synergizing reasoning and acting in language models}.
\newblock \emph{Preprint}, arXiv:2210.03629.

\bibitem[{Zhang et~al.(2022)Zhang, Zhang, Li, and Smola}]{zhang2022automatic}
Zhuosheng Zhang, Aston Zhang, Mu~Li, and Alex Smola. 2022.
\newblock \href {https://arxiv.org/abs/2210.03493} {Automatic chain of thought prompting in large language models}.
\newblock \emph{Preprint}, arXiv:2210.03493.

\bibitem[{Zhao et~al.(2021)Zhao, Wallace, Feng, Klein, and Singh}]{zhao2021calibrate}
Tony~Z. Zhao, Eric Wallace, Shi Feng, Dan Klein, and Sameer Singh. 2021.
\newblock \href {https://arxiv.org/abs/2102.09690} {Calibrate before use: Improving few-shot performance of language models}.
\newblock \emph{Preprint}, arXiv:2102.09690.

\bibitem[{Zheng et~al.(2023)Zheng, Xia, Zou, Dong, Wang, Xue, Wang, Shen, Wang, Li, Su, Yang, and Tang}]{zheng2023codegeex}
Qinkai Zheng, Xiao Xia, Xu~Zou, Yuxiao Dong, Shan Wang, Yufei Xue, Zihan Wang, Lei Shen, Andi Wang, Yang Li, Teng Su, Zhilin Yang, and Jie Tang. 2023.
\newblock \href {https://arxiv.org/abs/2303.17568} {Codegeex: A pre-trained model for code generation with multilingual evaluations on humaneval-x}.
\newblock \emph{Preprint}, arXiv:2303.17568.

\end{thebibliography}

\appendix

\section{Related Work}
\label{relatedwork}

\paragraph{Prompt Engineering}
The field of prompt engineering has garnered significant interest in recent years, in parts due to the emergent capabilities of the most capable LLMs, while also trying to better control their still unpredictable outcomes. A prominent strand of research within this domain concentrates on innovative prompting methodologies. These include few-shot prompting (\cite{brown2020language}), which enables models to adapt to new tasks without extensive retraining, and Chain-of-Thought prompting (\cite{wei2023chainofthought}), both of which are designed to enhance the reasoning capabilities of LLMs. Additionally, Automatic Chain-of-Thought (Auto-CoT) (\cite{zhang2022automatic}) and Self-Consistency (\cite{wang2023selfconsistency}) approaches have been developed to further refine these reasoning processes. To mitigate hallucinations in LLM outputs, techniques such as Retrieval Augmented Generation (RAG) (\cite{lewis2021retrievalaugmented}) and ReAct (\cite{yao2023react}) have been introduced. A thorough examination of these methodologies can be found in the survey by \cite{sahoo2024systematic}. In recent developments, a novel prompt programming framework (\cite{guidance}) has been introduced, which offers greater control and efficiency in generating structured outputs. Our study diverges from this approach by examining the effects of more prevalent and established prompt formats on LLMs, as opposed to investigating formats that are newly proposed and not widely adopted yet. Furthermore, it is important to note that third-party tools are predominantly designed for integration with open-source models, which may not seamlessly extend to proprietary models such as GPT. Another similar vein of research is dedicated to the structural design of prompts, aiming to optimize task performance without altering the inherent semantic content. This includes investigations into the sequential arrangement of context (\cite{liu2023lost, zhao2021calibrate, lu2022fantastically}) and the design of prompt formats (\cite{sclar2023quantifying, voronov2024mind, shu2023you}). Our work contributes to this growing body of literature by examining the impact of prompt formatting on the performance of LLMs.

\paragraph{Prompt Format}
The sensitivity of LLMs to prompt construction is a well-documented phenomenon, yet research on the impact of prompt formats on model performance remains sparse. Pioneering studies ( \cite{sclar2023quantifying, voronov2024mind, shu2023you}) have conducted rigorous investigations, revealing that widely used open-source LLMs exhibit extreme sensitivity to variations in prompt format. These studies, however, primarily focus on subtle, local changes to the format—such as the number of colons following a question, the insertion of newlines, or the selection of input/output verbalizers. Besides, their experimental designs are confined to classification tasks, limiting the generalizability of findings across diverse tasks.

Our research diverges from these existing studies by examining the effects of global prompt format modifications on model performance, offering insights that are applicable to a broad spectrum of LLM-based tasks that necessitate prompt engineering. The closest related work to ours is by \cite{sui2024table}, which however only provides a cursory exploration of format influence and is restricted to tabular data. To the best of our knowledge, our study is the first effort to systematically investigate the impact of global prompt format variations - an inescapable aspect of prompt engineering design decisions.

\section{Datasets}
\label{appendix:datadescription}
We evaluate six distinct benchmarks and classify them according to the nature of the task involved.
\paragraph{NL2NL}
\begin{itemize}
    \item \textbf{Massive Multitask Language Understanding (MMLU)} \cite{hendrycks2020measuring} covers 57 subjects including 20 STEM subjects, 13 humanities subjects, 12 social sciences subjects and 12 other subjects. Each subject contains at least 100 multiple choice questions, which tests both world knowledge and problem solving ability. We use the dev set which contains 5 questions per subjects as few-shot examples, and use test set containing 14,079 questions with different levels of difficulty to evaluation model performance. We use accuracy score to measure model performance.

    \item \textbf{NER Finance:} OpenAI Evals \cite{openai_evals} is a framework containing a registry of evaluations to test LLMs where NER Finance is one of those. This dataset contains samples between one sentence to one paragraph long from financial documents. The task is to extract the all of the entities in the document, with the evaluation being if the LLM outputs each entity, in order. We randomly sample 500 examples from this dataset.
\end{itemize}

\paragraph{NL2Code}
    \begin{itemize}
        \item \textbf{HumanEval} \cite{chen2021codex} is a benchmark dataset consisting of a collection of Python programming problems, each accompanied by a function signature, a docstring outlining the problem to be solved, and a set of unit tests that the correct implementation must pass. We use the evaluation metric pass@1, which checks if the the generated code passes the unit given unit tests in one attempt. We use all 164 samples in this dataset. 
        \item \textbf{FIND} \cite{schwettmann2023find}: The Function Interpretation and Description (FIND) benchmark dataset is a natural language-to-code generation task. The LLM is given $5$ example inputs and outputs to an unknown Python function and is tasked with reverse engineering the original Python code. We evaluate the benchmark by comparing the output of test cases on a ground truth function with the output from LLM generated functions. We use the "strings" category of functions for the task consisting of 500 functions.  We provide the LLM with $5$ pairs of input and output for each function. To select these examples, we randomly sample from a dataset provided by \cite{schwettmann2023find} that contains example test strings for each function. To evaluate the generated function code, we use the string indicator metric introduced by \cite{schwettmann2023find} that measures the number of test cases passed by the function.  
    \end{itemize}
\paragraph{Code2Code}
    \begin{itemize}
        \item \textbf{CODEXGLUE} \cite{lu2021codexglue} stands for General Language Understanding Evaluation benchmark for CODE. It was originally introduced to address the lack of diversified benchmarks in code intelligence by providing a diverse set of tasks, including code translation. We use the parallel code for Java and C-Sharp and vice versa. We use the test set containing 1000 parallel code in Java and C-Sharp to experiment the capabilities of the LLMs in translating code from one programming language to another. The performance of the LLMs is evaluated using the BLEU \cite{papineni2002bleu} score, which compares the generated code to the reference code.
        \item \textbf{HumanEval-X} \cite{zheng2023codegeex} dataset is a benchmark designed to evaluate the multilingual capabilities of code generative models. It contains 820 high-quality, human-crafted data samples, each accompanied by test cases. The dataset supports a variety of programming languages, including Python, C++, Java, JavaScript, and Go. In this we experiment with one of the above dimension of code-translation focusing on Java to Python. To accomplish this task, we combine the "declaration" and "canonical-solution" together to finally get the overall function in the respective language. "Declaration" contains the function declaration for the respective language and "canonical solution" has the human-crafted example solution for the language. Similar to CODEXGLUE, we use the BLEU  \cite{papineni2002bleu} score for measuring the performance.
    \end{itemize}

\begin{table*}[!ht]
\centering
\scalebox{0.9}{
\begin{tabular}{|c|c|}
\hline

\textbf{Prompt Format} & \textbf{Prompt Template}\\
\hline
Plain text &  \makecell[l]{ \{persona\} \{instructions\} \{examples\} \\ \{output format instructions\} \{user ask\} } \\
\hline
Markdown  &  \makecell[l]{ \#\# Persona\\ \{persona\} \\ \#\# Instructions \\ \{instructions\} \\ \#\# Examples \\ \{examples\} \\ \#\# Output Format \\ \{Output format instructions\} \\ \#\# User Question \\ \{user ask\} } \\
\hline
YAML  & \makecell[l]{ Persona\\ - \{persona\} \\ Instructions \\ - \{instructions\} \\ Examples \\ - \{examples\} \\ Output format \\ - \{output format instructions\} \\ User question \\ - \{user ask\} } \\
\hline
JSON & \makecell[l]{\{\\ \hspace*{2em}``Persona'': ``\{persona\}'',\\  \hspace*{2em}``Instructions'': ''\{instructions\}``,\\ \hspace*{2em}``Examples'': ``\{examples\}'',\\ \hspace*{2em}``Output format'': ``\{output format instructions\}" \\ \} \\ \{\\ \hspace*{2em}``User ask'': ``\{user ask\}'' \\ \}} \\
\hline
\end{tabular}
}
\caption{\small Prompt templates considered in this paper. Placeholders are denoted with \{variable name\} and get replaced with task specific context.}
\label{table:promptformatillustration}
\end{table*}

\section{Prompt Templates}
\label{appendix:prompt_template}
In this section we provide examples of the four prompt templates we used for the NER Finance task. Prompts for all other tasks followed identical formatting. Variables that are injected into the prompt are denoted by blue text wrapped in braces. For example a user ask being injected is denoted as \textcolor{blue}{\{USER ASK\}}.

\begin{figure*}[!ht]
 \centering
    \begin{AIbox}{Plaintext}
        \parbox[t]{0.99\textwidth}{
            \begin{alltt}
            \scriptsize
                \textbf{System:}\\
                    You are a annotator working for large financial data company and are tasked with extracting named entities from financial documents who follows strict guidelines for quality and formatting. The following sentence is from a financial document. List the named entities in the order they appear in the sentence. If an entity appears multiple times, list it multiples times. Entities should be stated in the format NAME - TYPE where TYPE can be PERSON, ORGANIZATION, or LOCATION. State your final answer as a comma-separated list of entities enclosed in square brackets. Example: [Bank - ORGANIZATION, Borrower - PERSON]. If there are no entities found, state your final answer as 'No entities found'. Provide your chain of thought first and then respond with your final answer. Here is an example: \textcolor{blue}{\{ICL EXAMPLE INPUT\}} \textcolor{blue}{\{ICL EXAMPLE SOLUTION\}}\\

                \textbf{User:}\\
                \textcolor{blue}{\{INPUT\}}\\
            \end{alltt}}
    \end{AIbox}
    \label{fig:prompt_answer_pt}
\end{figure*}

\begin{figure*}[!ht]
    \begin{AIbox}{Markdown}
         
        \parbox[t]{0.99\textwidth}{
            \begin{alltt}
            \scriptsize
                \textbf{System:}\\
                    \#\# Persona\\
                    - You are a annotator working for large financial data company are tasked with extracting named entities from financial documents who follows strict guidelines for quality and formatting. \\
                    \\
                    \#\# Instructions\\
                    - You will be given a sentence from a financial document. 
                    - List the named entities in the order they appear in the sentence.\\
                    - If an entity appears multiple times, list it multiples times.\\
                    - Provide your chain of thought first and then respond with your final answer.\\
                    \\
                    \#\# Output Format\\
                    - Entities should be stated in the format NAME - TYPE where TYPE can be PERSON, ORGANIZATION, or LOCATION.\\
                    - State your final answer as a comma-separated list of entities enclosed in square brackets. Example: [Bank - ORGANIZATION, Borrower - PERSON].\\
                    - If there are no entities found, state your final answer as 'No entities found'.\\
                    \\
                    \#\# Example\\
                    \#\#\# DOCUMENT\\
                    \textcolor{blue}{\{ICL EXAMPLE INPUT\}}\\
                    \\
                    \#\#\# Solution\\
                    \textcolor{blue}{\{ICL EXAMPLE SOLUTION\}}\\
                    
                \textbf{User:}\\
                \#\#\# DOCUMENT\\
                \textcolor{blue}{\{INPUT\}}\\
            \end{alltt}}
    \end{AIbox}
    \label{fig:prompt_answer_md}
\end{figure*}

\begin{figure*}[!ht]
    \begin{AIbox}{YAML}
         
        \parbox[t]{0.99\textwidth}{
            \begin{alltt}
            \scriptsize
                \textbf{System:}\\
                    Persona:\\
                  \hspace*{0.25cm}Description: You are a annotator working for large financial data company are tasked with extracting named entities from financial documents who follows strict guidelines for quality and formatting.\\
                
                Instructions:\\
                  \hspace*{0.25cm}- You will be given a sentence from a financial document. \\
                  \hspace*{0.25cm}- List the named entities in the order they appear in the sentence.\\
                  \hspace*{0.25cm}- If an entity appears multiple times, list it multiples times.\\
                  \hspace*{0.25cm}- Provide your chain of thought first and then respond with your final answer.\\
                
                Output\_Format:\\
                  \hspace*{0.25cm}Entities should be stated in the format NAME - TYPE where TYPE can be PERSON, ORGANIZATION, or LOCATION. State your final answer as a comma-separated list of entities enclosed in square brackets. \\
                
                Examples:\\
                 \hspace*{0.25cm}- Document: \textcolor{blue}{\{ICL EXAMPLE INPUT\}}\\
                 \hspace*{0.25cm}- Solution: \textcolor{blue}{\{ICL EXAMPLE SOLUTION\}}\\
 
                \textbf{User:}\\
                Task:\\
                \hspace*{0.25cm}- Document: \textcolor{blue}{\{INPUT\}}\\
                
            \end{alltt}}
    \end{AIbox}
    \label{fig:prompt_answer_yaml}
\end{figure*}

\begin{figure*}[!ht]
    \begin{AIbox}{JSON}
        \parbox[t]{0.99\textwidth}{
            \begin{alltt}
            \scriptsize
                \textbf{System:}\\
                   \{\\
                        \hspace*{0.25cm}"Persona": "You are a annotator working for large financial data company are tasked with extracting named entities from financial documents who follows strict guidelines for quality and formatting.",
                        
                        \hspace*{0.25cm}"Instructions": [\\
                            \hspace*{0.5cm}"You will be given a sentence from a financial document.",\\
                            \hspace*{0.5cm}"List the named entities in the order they appear in the sentence.",\\
                            \hspace*{0.5cm}"If an entity appears multiple times, list it multiples times.",\\
                            \hspace*{0.5cm}"Provide your chain of thought first and then respond with your final answer."\\
                        \hspace*{0.25cm}],
                    
                        \hspace*{0.25cm}"OutputFormat": "Entities should be stated in the format NAME - TYPE where TYPE can be PERSON, ORGANIZATION, or LOCATION. State your final answer as a comma-separated list of entities enclosed in square brackets. Example: [Bank - ORGANIZATION, Borrower - PERSON]. If there are no entities found, state your final answer as 'No entities found'.",
                    
                        \hspace*{0.25cm}"Example": "\textcolor{blue}{\{ICL EXAMPLE INPUT\}}\textbackslash n\textcolor{blue}{\{ICL EXAMPLE SOLUTION\}}"\\
                    \}\\
 
                \textbf{User:}\\
                \{\\
                    \hspace*{0.25cm}"Task": "\textcolor{blue}{\{INPUT\}}"\\
                \}\\
                
            \end{alltt}}
    \end{AIbox}
    \label{fig:prompt_answer_json}
\end{figure*}

\section{Additional Research Questions}

\begin{figure*}[!ht]
 \centering
    \begin{subfigure}{.33\linewidth}
        \centering
        \includegraphics[width=\linewidth]{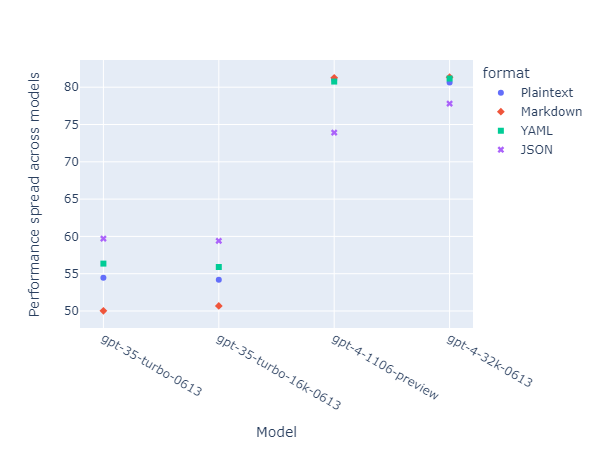}
        \caption{MMLU}
        \label{sfig:mmlu}
    \end{subfigure}%
    \hfill
    \begin{subfigure}{.33\linewidth}
        \centering
        \includegraphics[width=\linewidth]{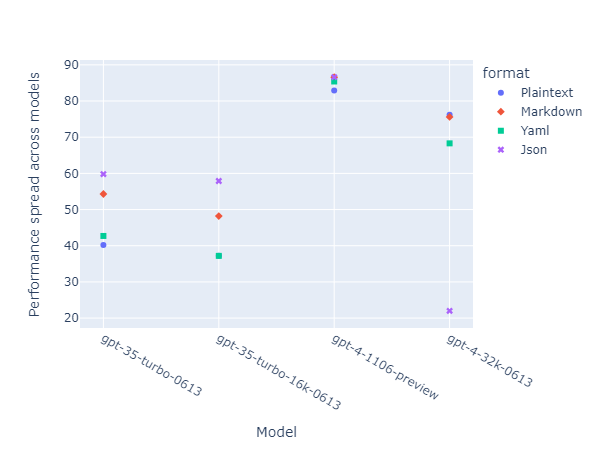}
        \caption{HumanEval}
        \label{sfig:HumanEval}
    \end{subfigure}%
    \hfill
    \begin{subfigure}{.33\linewidth}
        \centering
        \includegraphics[width=\linewidth]{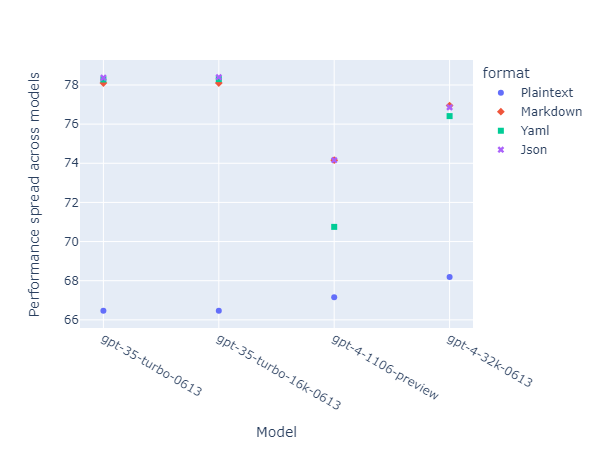}
        \caption{CODEXGLUE (Java2CS)}
        \label{sfig:java2cs}
    \end{subfigure}
\caption{ Model performance across prompt formats on MMLU, HumanEval and CODEXGLUE. Performance measurement for MMLU is accuracy, for HumanEval is pass@1 to hecks if the the generated code passes the
unit given unit tests in one attempt, for CODEXGLUE(Java2CS) is BLEU score. Plots for the remaining datasets are in Figure \ref{fig:dotplot_all}.}
\label{fig:dotplot}
\end{figure*}

\subsection{How does the format in which information is structured and presented influence the ability to solve problems that require different skill sets?}

We analyze whether model's sensitivity to prompt format changes is related to the skills required to solve the task using the MMLU benchmark which comprises 57 subjects, categorized into four domains: humanities, social science, STEM, and others. Each domain encompasses various disciplines, necessitating distinct skill sets and knowledge for accurate question answering. 

\begin{figure*}[!ht]
\center
  \includegraphics[width=16cm, height=7cm]{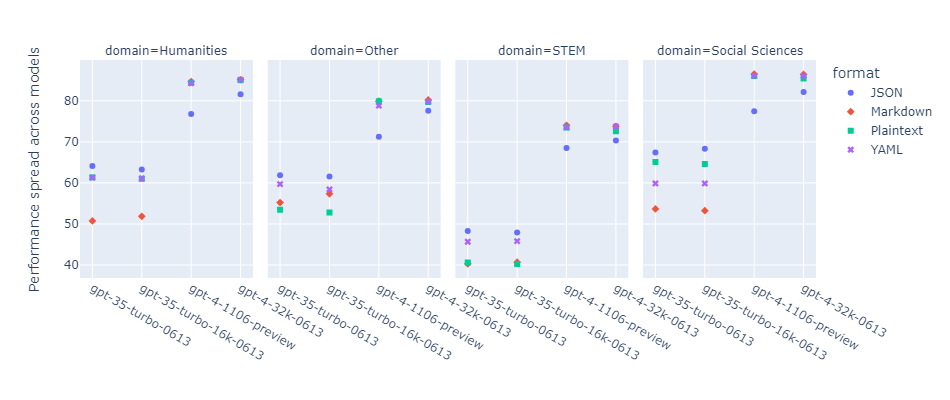}
\caption{  Performance spread across models on MMLU benchmark per domain. Wide performance spread is observed across domains that required different skills.}
\label{fig:mmlubreakdowndotplot}
\end{figure*}

Figure \ref{fig:mmlubreakdowndotplot} breaks down the performance on MMLU dataset by domain. We observe the performance spread exists across different tasks, and it's not signified nor eliminated by specific skills required. This suggests that the model’s sensitivity to prompt formatting is a general characteristic, rather than being contingent on the specific skills or reasoning abilities required by different tasks. Performance is influenced by how information is presented to it, regardless of the complexity or nature of the task, the way in which a problem is framed and communicated to the model can significantly impact its ability to process and respond to the information. \textbf{Model performance is consistently influenced by prompt formatting across various tasks, regardless of the specific skills or knowledge required.}

\subsection{Is there a correlation between the model size and the robustness of the LLM for different prompt templates?}

\begin{figure}[!ht]
\center
  \includegraphics[height=4.5cm, width=5.5cm]{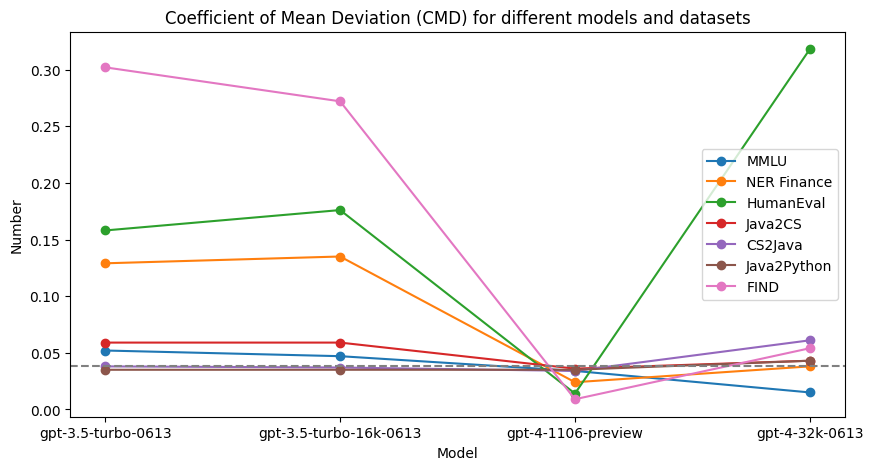}
\caption{Coefficient of mean deviation (CMD) of scalar metrics for all the prompt templates. Figure shows the CMDs across models and datasets. GPT-3.5 series exhibit larger CMD scores across benchmarks than GPT-4 series, indicating higher sensitivity to the choice of format. }
\label{fig:cmd}
\end{figure}

\paragraph{Coefficient of Mean Deviation.} We measure the performance dispersion for a model caused by format and compare if the degree of dispersion can be attributed to the size of the model. We compute the Coefficient of Mean Deviation (CMD) across all the prompt templates for every model.
$$ CMD = \frac{\sum{|s(p_i) - \bar{s}|}}{n \cdot \bar{s}} $$ \\
Here \( s(p_i) \) is the performance score for prompt $p_i$, \( \bar{s} \) is the average score across all prompt formats, and \( n \) is the number of formats.  A lower CMD indicates that the model's performance is less affected by prompt format changes, showing greater robustness. A higher CMD suggests more variability and sensitivity to prompt format.

While the architectural details and exact size of GPT-4 are not published, it is assumed that GPT-4 contains significantly more parameters, was trained on more data than GPT-3.5, and is clearly the overall more capable model (\cite{achiam2023gpt, bubeck2023sparks, carlini2024stealing}). In this section, we aim to ascertain whether an expansion in general capability translates to enhanced stability in response to changes in templates. The CMDs for all the models across benchmarks are presented in Figure \ref{fig:cmd}.

A lower value of CMD indicates more robustness to template variation. The results indicate that the GPT-4-1106-preview model exhibits superior robustness to format changes, maintaining a performance dispersion consistently below 0.036 across all benchmarks. In contrast, the GPT-4-32k-0613 model demonstrates less robustness relative to the GPT-4-1106-preview, yet it outperforms the GPT-3.5 series, with CMDs not exceeding 0.043. The GPT-3.5 series displays a broader range of CMDs, from 0.035 to 0.176, signifying a higher degree of performance variability under different prompt formats. GPT-4's observed improvements may be attributed to its enhanced ability to process data in diverse formats. Moreover, it is possible that the robustness of the model is not adversely impacted by format variations at the level of the last hidden layer of prompt embedding. Notably, the GPT-4-1106-preview model achieves greater robustness compared to the GPT-4-32k-0613, corroborating existing evidence that suggests the former has a heightened proficiency in comprehending and generating content in specific formats as instructed \cite{OpenAI}. Further examining GPT-4-32k-0613's performance, we notice the CMD on HumanEval benchmark is extremely high, this is due the extremely low score using JSON format, see Table \ref{table:human_eval_performance} for results. Analyzing the model outputs, we find the poor performance is because most of the time the model would generate chain of thought in plain text, but did not continue with actually generating the code. The other models did not exhibit this behavior for the JSON template. We hypothesize that this may be related to the OpenAI's claim about fixing laziness in task completion in the 0125 version of GPT-4-turbo \cite{OpenAI2024}. In summary, \textbf{larger models are more robust to template variation.}

\section{Complete Results}

\subsection{Additional results on model performance under all templates across benchamrks.}
\begin{table*}[!ht]
\centering
\begin{tabular}{lcccc}
\hline
\textbf{Model} & GPT-35-turbo-0613 & GPT-35-turbo-16k-0613 & GPT-4-1106-preview & GPT-4-32k-0613 \\
\hline
Plaintext & 54.464 $\pm$ 18.300& 54.184  $\pm$ 19.066& 81.005 $\pm$ 12.979& 80.638 $\pm$ 13.172\\
Markdown  & 50.021 $\pm$ 17.144 & 50.686 $\pm$ 17.436  & \textbf{81.252} $\pm$ 12.932& \textbf{81.349} $\pm$ 13.158\\
YAML  & 56.355 $\pm$ 16.792& 55.901 $\pm$ 16.347& 80.758 $\pm$ 13.000&  81.162 $\pm$ 13.110\\
JSON  & \textbf{59.705} $\pm$ 16.594& \textbf{59.405} $\pm$ 17.092&73.918 $\pm$ 13.580&  77.800 $\pm$ 13.725 \\
\hline
\end{tabular}
\caption{Model Performance on MMLU Benchmark. Accuracy is averaged over 57 different subjects. }
\label{table:mmluperformance}
\end{table*}

\begin{table*}[!ht]
\centering
\begin{tabular}{lcccc}
\hline
\textbf{Model} & GPT-35-turbo-0613 & GPT-35-turbo-16k-0613 & GPT-4-1106-preview & GPT-4-32k-0613 \\
\hline
Plaintext    & 40.24 $\pm$ 3.98 & 37.20 $\pm$ 3.77 & 82.93 $\pm$ 4.39 & \textbf{76.22} $\pm$ 4.76 \\
Markdown     & 54.27 $\pm$ 4.70 & 48.17 $\pm$ 4.44 & \textbf{86.59} $\pm$ 4.06 & 75.61 $\pm$ 4.78 \\
YAML         & 42.68 $\pm$ 4.14 & 37.20 $\pm$ 3.77 & 85.37 $\pm$ 4.18 & 68.29 $\pm$ 4.92 \\
JSON         & \textbf{59.76} $\pm$ 4.85 & \textbf{57.93} $\pm$ 4.81 & \textbf{86.59} $\pm$ 4.06 & 21.95 $\pm$ 2.48 \\
\hline
\end{tabular}
\caption{Model performance on HumanEval benchmark. We used all 164 samples for testing. }
\label{table:human_eval_performance}
\end{table*}

\begin{table*}[!ht]
\centering
\begin{tabular}{lcccc}
\hline
\textbf{Model} & GPT-35-turbo-0613 & GPT-35-turbo-16k-0613 & GPT-4-1106-preview & GPT-4-32k-0613 \\
\hline
Plaintext & \textbf{37.20} $\pm$ 6.59 & \textbf{36.80} $\pm$ 6.54 & 49.40 $\pm$ 7.86 & 47.20 $\pm$ 7.67 \\
Markdown  & 28.00 $\pm$ 5.31 & 30.00 $\pm$ 5.61 & 51.40 $\pm$ 8.01 & 51.60 $\pm$ 8.03 \\
YAML  & 24.60 $\pm$ 4.78 & 21.80 $\pm$ 4.31 & \textbf{53.80} $\pm$ 8.18 & \textbf{53.20} $\pm$ 8.14 \\
JSON  & 28.40 $\pm$ 5.37 & 31.00 $\pm$ 5.76 & 50.80 $\pm$ 7.97 & 52.40 $\pm$ 8.08 \\
\hline
\end{tabular}
\caption{Model performance on NER finance benchmark. We randomly sampled 500 samples for testing. }
\label{table:ner_finance_performance}
\end{table*}

\begin{table*}[!ht]
\centering
\begin{tabular}{lcccc}
\hline
\textbf{Model} & gpt-35-turbo-16k-0613 & gpt-35-turbo-0613 & gpt-4-32k-0613 & gpt-4-1106-preview\\ \hline    
plaintext & 15.75$\pm$0.142 & 15.9$\pm$0.143 & 21.87$\pm$0.164 & 20.08$\pm$0.161\\
markdown & 5.03$\pm$0.092 & 5.19$\pm$0.089 & 17.42$\pm$0.154 & 20.68$\pm$0.165\\
json & 14.46$\pm$0.138 & 14.33$\pm$0.144 & 21.15$\pm$0.162 & 20.19$\pm$0.156\\
yaml & 13.06$\pm$0.139 & 13.49$\pm$0.138 & 21.6$\pm$0.163 & 20.28$\pm$0.16\\ \hline
\end{tabular}
\caption{Model performance on FIND benchmark. Test set includes 500 functions.}
\label{table:find_model_performance}
\end{table*}

\begin{table*}[!ht]
\centering
\begin{tabular}{lcccc}
\hline
\textbf{Model} & GPT-35-turbo-0613 & GPT-35-turbo-16k-0613 & GPT-4-1106-preview & GPT-4-32k-0613  \\
\hline
Plaintext & 62.95 $\pm$ 15.64 & 62.92 $\pm$ 15.66 & 64.95 $\pm$ 15.52 & 63.86 $\pm$ 15.83 \\
Markdown  & 69.82 $\pm$ 16.26 & 69.84 $\pm$ 16.23 & 70.70 $\pm$ 17.44 & 71.65 $\pm$ 18.10 \\
YAML      & 69.05 $\pm$ 17.24 & 69.05 $\pm$ 17.24 & 71.16 $\pm$ 16.11 & 71.41 $\pm$ 17.96 \\
JSON      & 68.85 $\pm$ 16.13 & 68.85 $\pm$ 16.13 & 72.30 $\pm$ 15.97 & 72.39 $\pm$ 16.46 \\ 
\hline
\end{tabular}
\caption{Model performance on HumanEval-X, a Java to Python translation task The test set contains 164 data samples.}
\label{table:javatopythonconsistency}
\end{table*}

\begin{table*}[!ht]
\centering
\resizebox{\textwidth}{!}{%
\begin{tabular}{lcccc c cccc }
\hline
\textbf{Dataset} & \multicolumn{4}{c}{CODEXGLUE: Java to CS}   && \multicolumn{4}{c}{CODEXGLUE - CS to Java}\\
\cline{2-5} \cline{7-10}
\textbf{Model} & GPT-35-turbo-0613 & GPT-35-turbo-16k-0613 & GPT-4-1106-preview & GPT-4-32k-0613 &  & GPT-35-turbo-0613 & GPT-35-turbo-16k-0613 & GPT-4-1106-preview & GPT-4-32k-0613\\
\hline
Plaintext  & 66.46 $\pm$ 16.04 & 66.46 $\pm$ 16.04 & 67.16 $\pm$ 16.77 & 68.19 $\pm$ 13.14 &   & 68.81 $\pm$ 17.65 & 68.89 $\pm$ 17.64 & 67.93 $\pm$ 17.72 & 68.11 $\pm$ 16.29  \\
Markdown   & 78.10 $\pm$ 18.75 & 78.10 $\pm$ 18.75 & 74.16 $\pm$ 16.77 & 76.95 $\pm$ 18.33 &   &  76.19 $\pm$ 18.40 & 76.12 $\pm$ 18.37 & 74.80 $\pm$ 20.14 & 80.36 $\pm$ 20.52  \\
YAML       & 78.28 $\pm$ 18.92 & 78.30 $\pm$ 18.92 & 70.75 $\pm$ 16.08 & 76.41 $\pm$ 18.00 &   &  75.47 $\pm$ 20.16 & 75.39 $\pm$ 20.08 & 72.09 $\pm$ 17.98 & 78.49 $\pm$ 21.23  \\
JSON       & 78.37  $\pm$ 18.93 & 78.40  $\pm$ 18.93 & 74.16  $\pm$ 16.77 & 76.86  $\pm$ 18.31 &   &  77.49  $\pm$ 19.51 & 77.49  $\pm$ 19.50 & 75.00  $\pm$ 17.66 & 83.05  $\pm$ 18.60 \\
\hline
\end{tabular}
}
\caption{Model performance on Java to C\# and C\# to Java translation task. The test set contains 1000 samples in Java and C\#.}
\label{table:model_consistency}
\end{table*}

\subsection{IoU scores on all benchmarks.}

\begin{figure*}[!ht]
\centering
\begin{subfigure}{.4\linewidth}
  \centering
  \includegraphics[width=\linewidth]{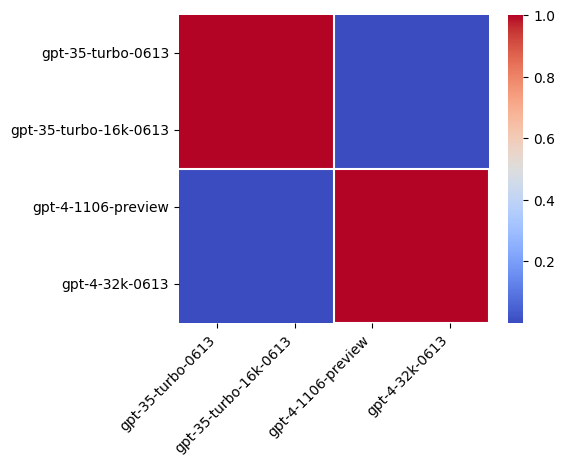}
  \caption{MMLU}
  \label{sfig:mmlu_heatmap}
\end{subfigure}\hfill
\begin{subfigure}{.4\linewidth}
  \centering
  \includegraphics[width=\linewidth]{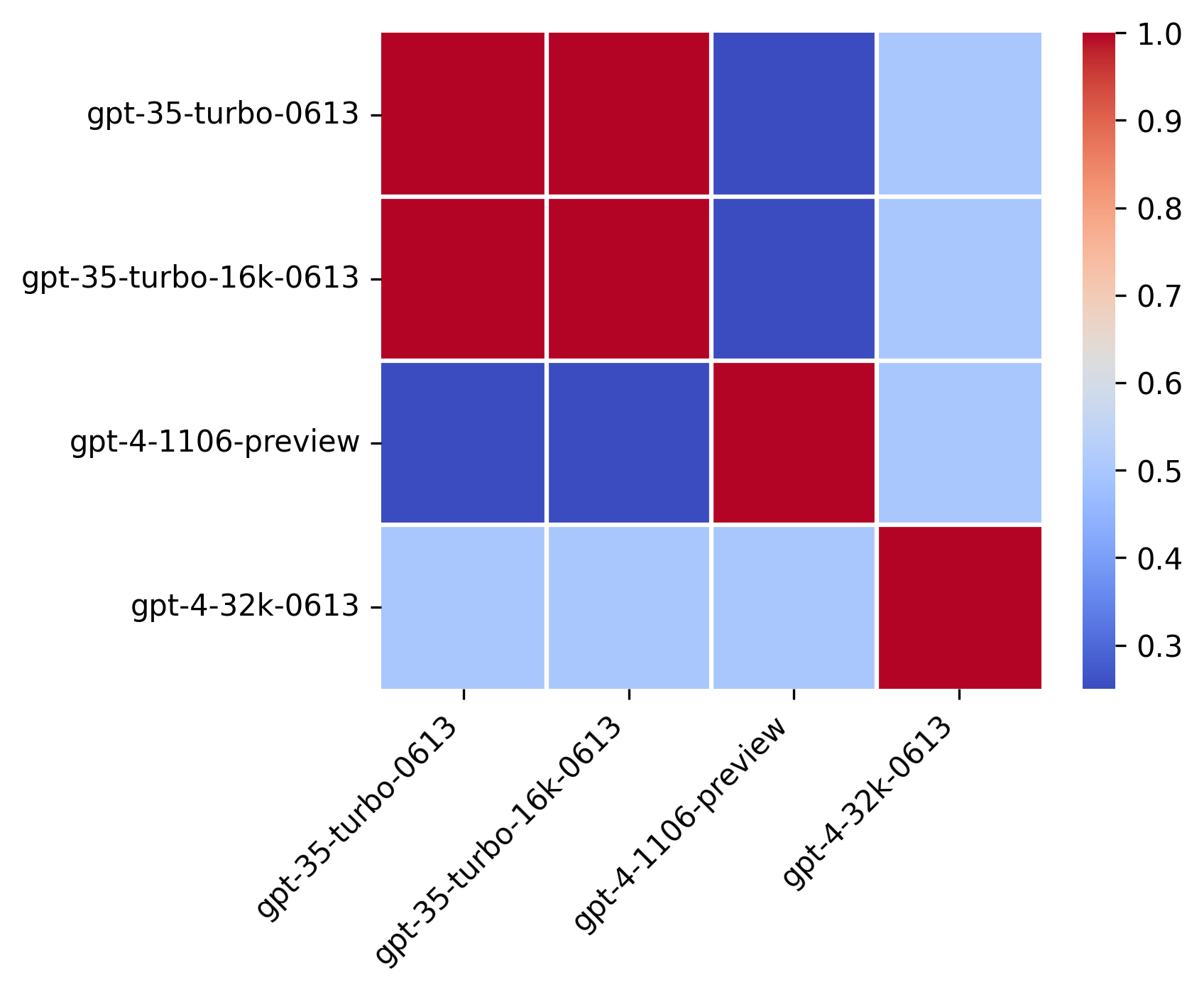}
  \caption{FIND}
  \label{sfig:find}
\end{subfigure}
\caption{Heatmap of IoU values for other benchmarks.}
\label{fig:IoUAll}
\end{figure*}

\begin{figure}[hbt!]
  \centering
  \begin{subfigure}{0.24\textwidth}
    \centering
    \includegraphics[width=\linewidth]{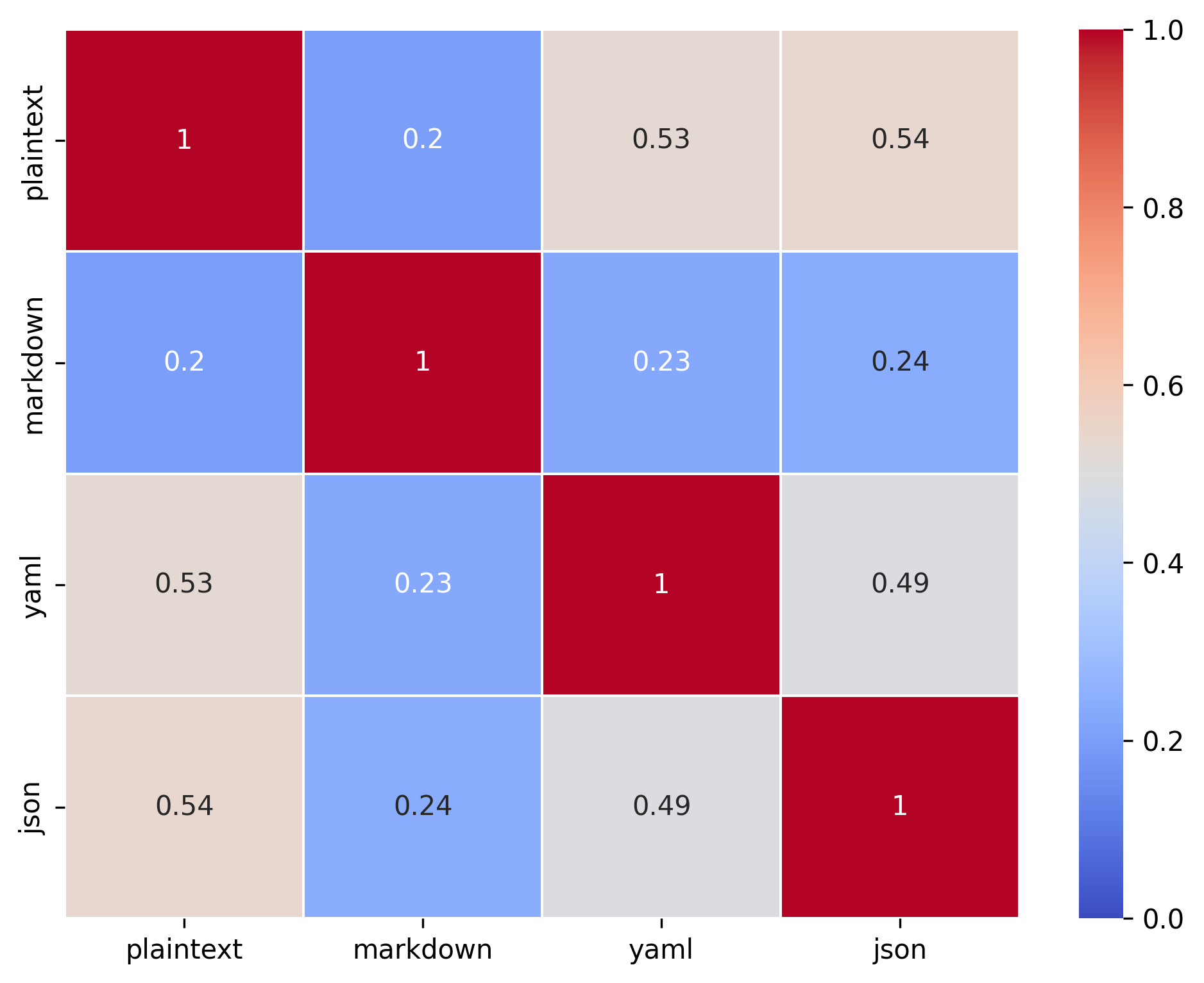}
    \caption{\fontsize{7}{9}\selectfont GPT-35-Turbo-0613}
    \label{fig:consistency-find-gpt35}
  \end{subfigure}%
  \hfill
  \begin{subfigure}{0.24\textwidth}
    \centering
    \includegraphics[width=\linewidth]{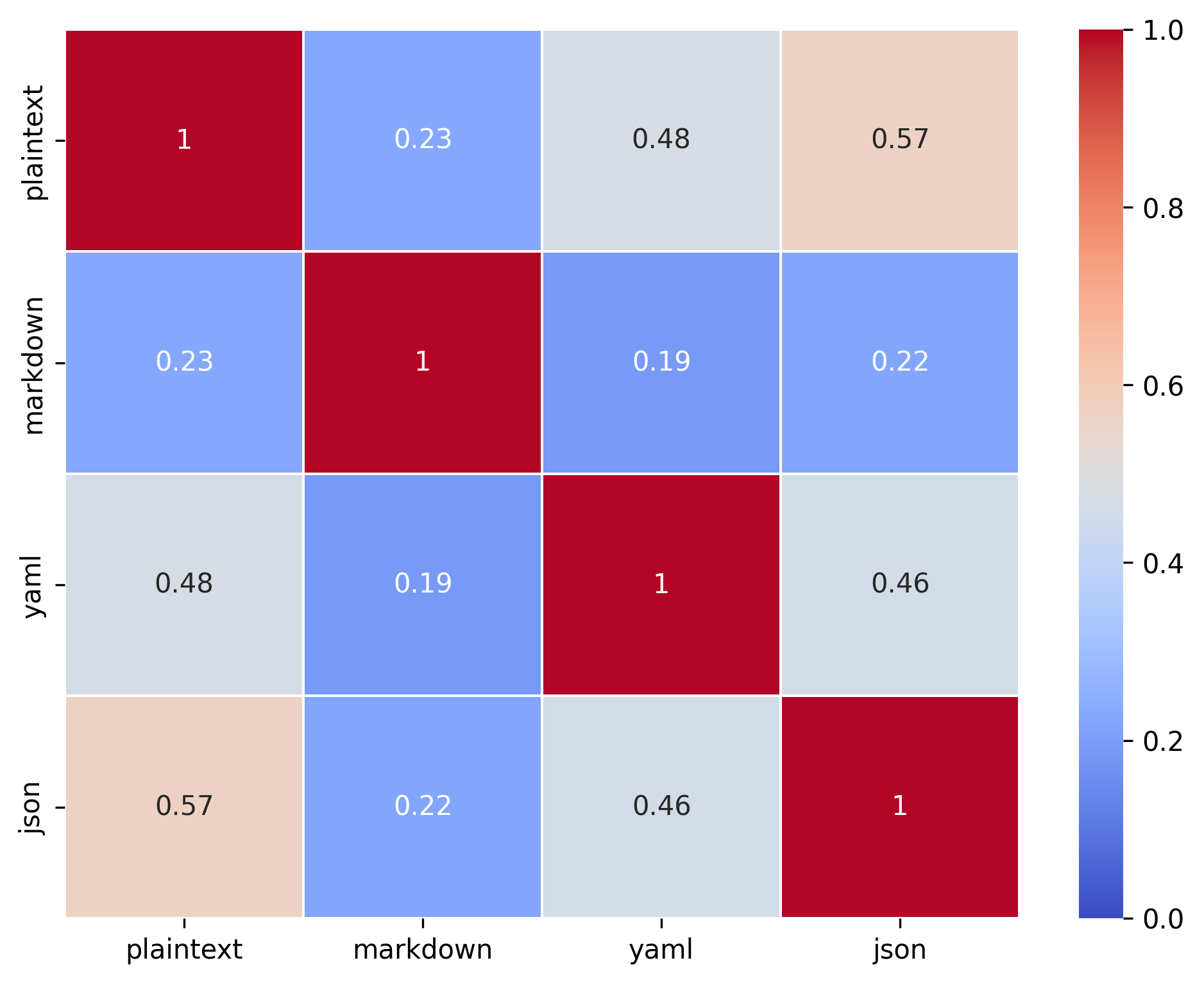}
    \caption{\fontsize{7}{9}\selectfont GPT-35-turbo-16k-0613}
    \label{fig:consistency-find-gpt35-16k}
  \end{subfigure}%
  \hfill
  \begin{subfigure}{0.24\textwidth}
    \centering
    \includegraphics[width=\linewidth]{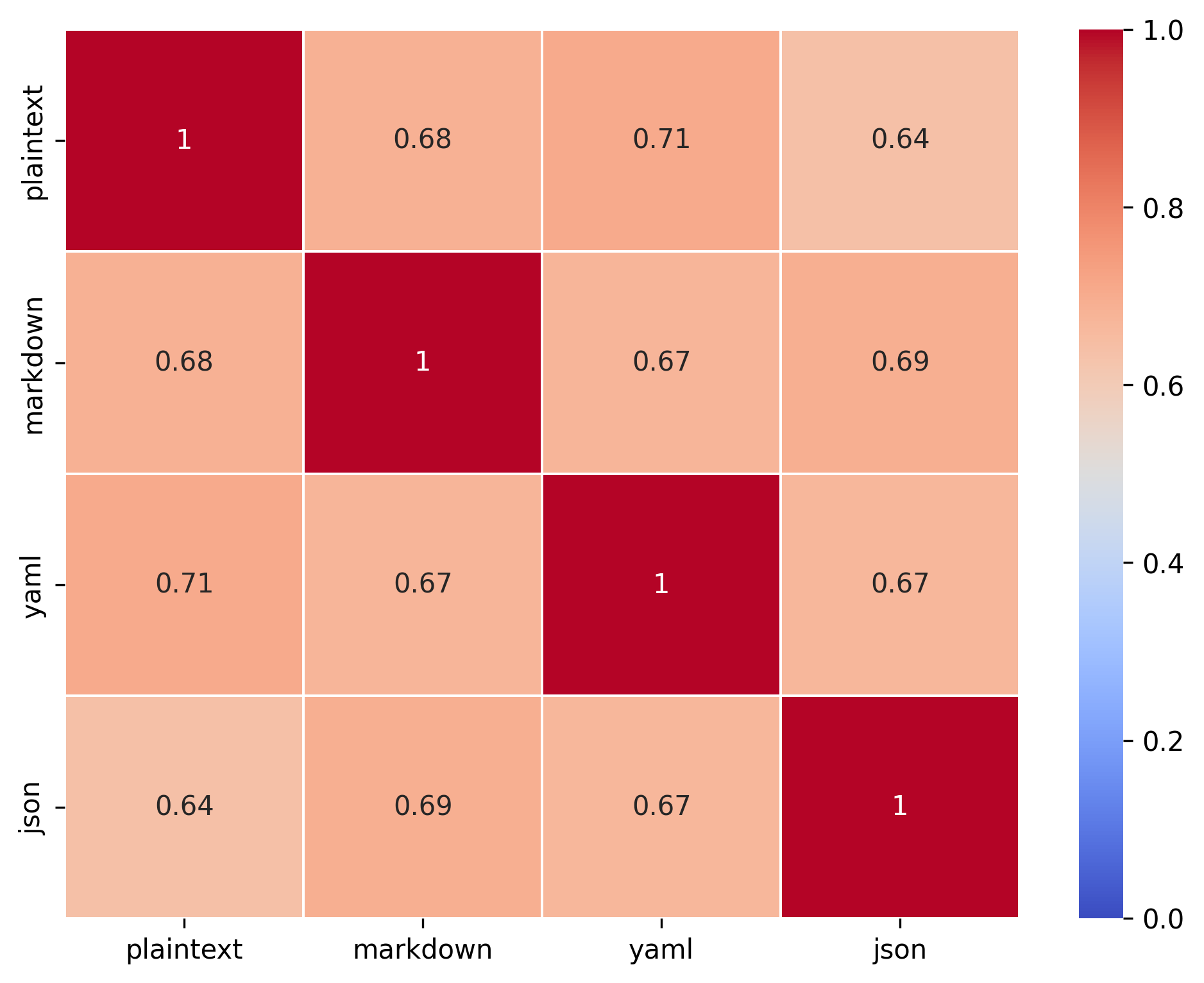}
    \caption{\fontsize{7}{9}\selectfont GPT-4-1106-preview}
    \label{fig:consistency-find-gpt4}
  \end{subfigure}%
  \hfill
  \begin{subfigure}{0.24\textwidth}
    \centering
    \includegraphics[width=\linewidth]{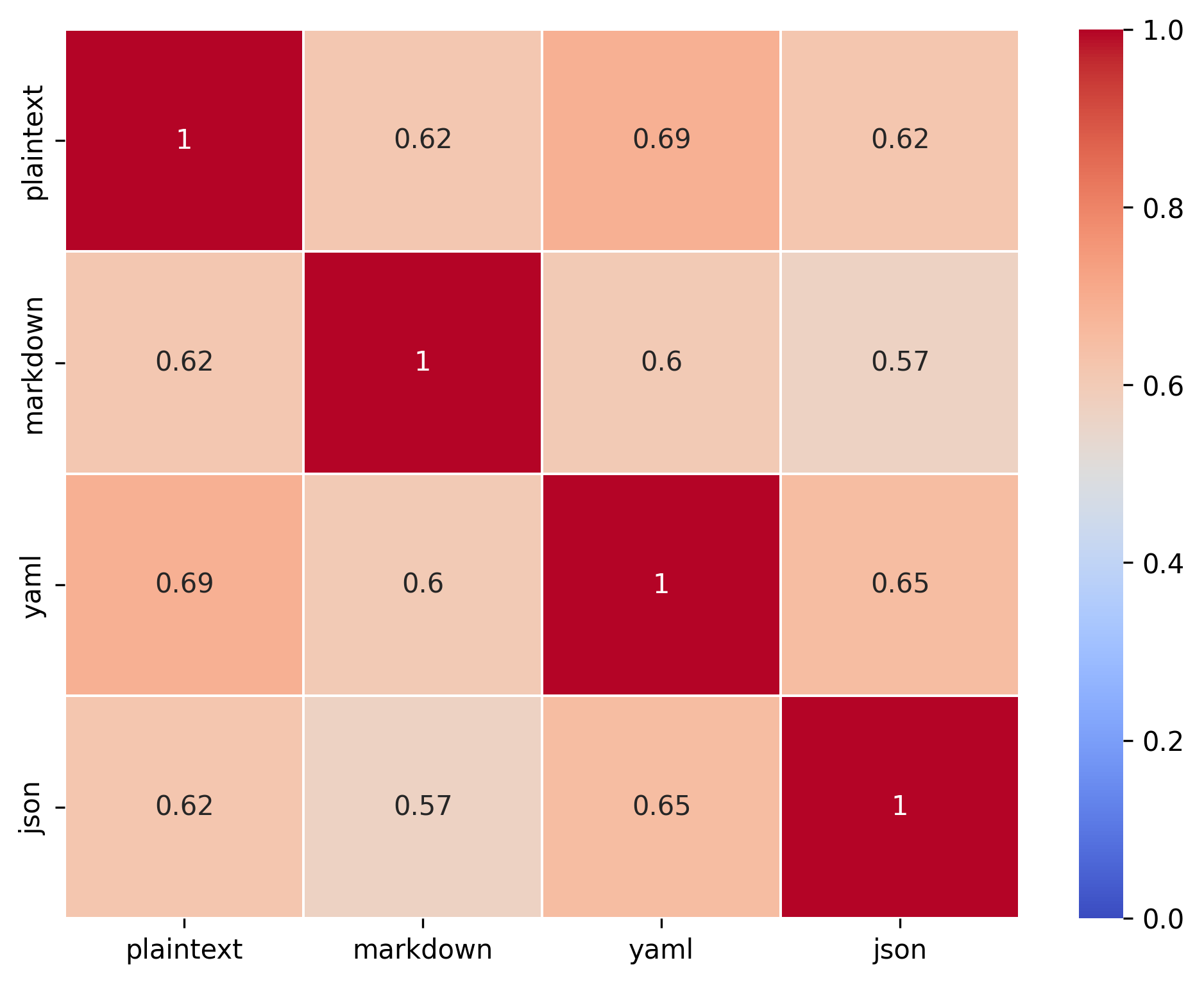}
    \caption{\fontsize{7}{9}\selectfont GPT-4-32k-0613}
    \label{fig:consistency-find-gpt4-32k}
  \end{subfigure}
  \caption{Performance of Consistency for FIND dataset across models.  }
  \label{fig:consistency-find}
\end{figure}

\subsection{Dotplot on all benchmark datasets}

\begin{figure*}[!ht]
  \centering
  \begin{subfigure}[b]{0.41\textwidth}
    \includegraphics[width=\linewidth]{latex/figures/mmludotplotmodelformat.png}
    \caption{MMLU}
    \label{sfig:mmlu_dotplot}
  \end{subfigure}
  \hfill
  \begin{subfigure}[b]{0.41\textwidth}
    \includegraphics[width=\linewidth]{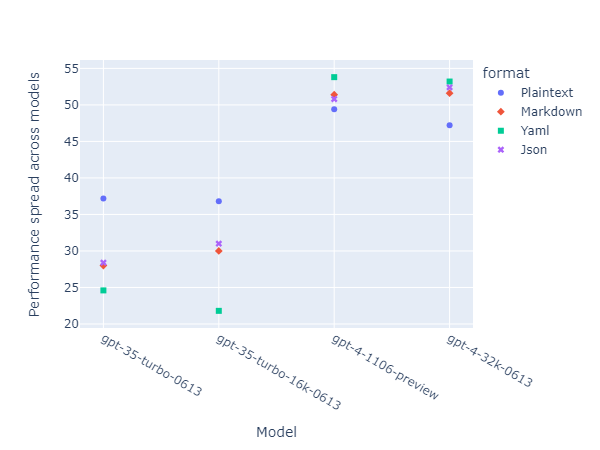}
    \caption{NER Finance}
    \label{sfig:ner}
  \end{subfigure}
  
  
  \begin{subfigure}[b]{0.41\textwidth}
    \includegraphics[width=\linewidth]{latex/figures/java2cs.png}
    \caption{CODEXGLUE (Java2CS)}
    \label{sfig:java2cs_dotplot}
  \end{subfigure}
  \hfill
  \begin{subfigure}[b]{0.41\textwidth}
    \includegraphics[width=\linewidth]{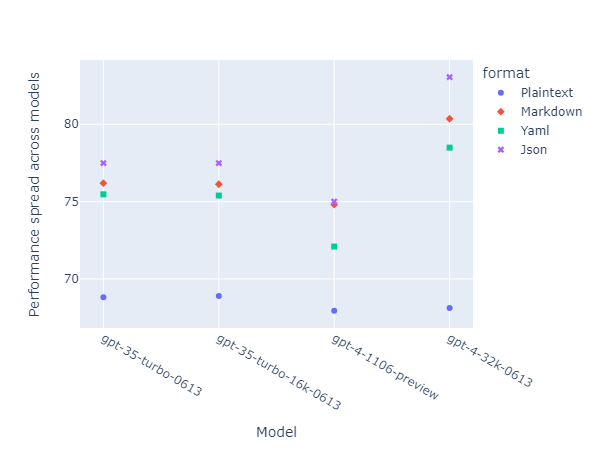}
    \caption{CODEXGLUE (CS2Java)}
    \label{sfig:cs2java}
  \end{subfigure}
  
  
  \begin{subfigure}[b]{0.41\textwidth}
    \includegraphics[width=\linewidth]{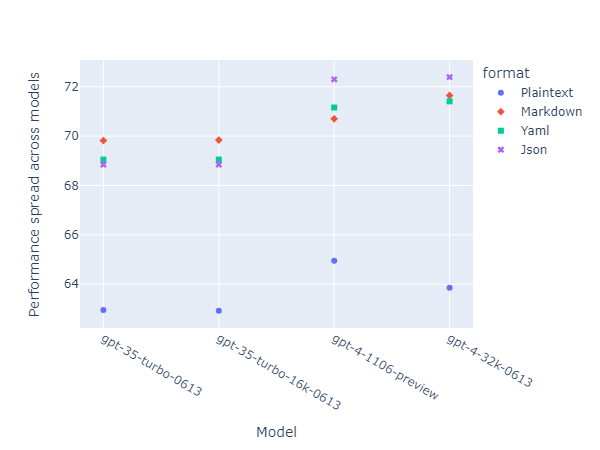}
    \caption{HumanEval-X (Java2Python)}
    \label{sfig:java2python}
  \end{subfigure}
  \hfill
  \begin{subfigure}[b]{0.41\textwidth}
    \includegraphics[width=\linewidth]{latex/figures/humanevaldotplot.png}
    \caption{HumanEval}
    \label{sfig:HumanEval_dotplot}
  \end{subfigure}
  
  \vspace{1ex} 
  
  \begin{subfigure}[b]{0.41\textwidth}
    \includegraphics[width=\linewidth]{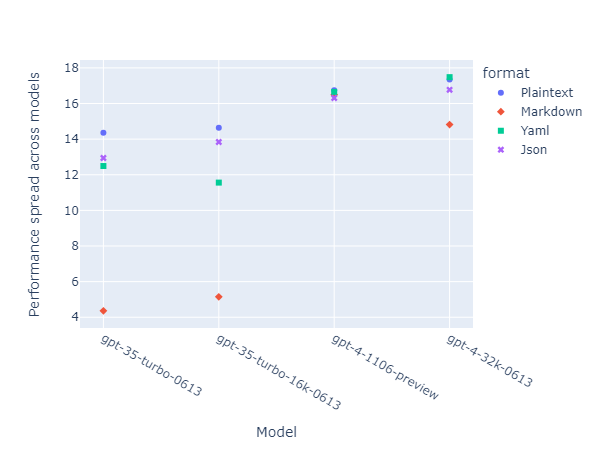}
    \caption{FIND}
    \label{sfig:FIND}
  \end{subfigure}
  
  \caption{Dotplot of model performance across prompt formats on all benchmarks.}
  \label{fig:dotplot_all}
\end{figure*}


\end{document}